\documentclass[10pt,twocolumn,letterpaper]{article}

\usepackage{wacv}
\usepackage{times}
\usepackage{epsfig}
\usepackage{graphicx}
\usepackage{amsmath}
\usepackage{amssymb}
\usepackage{subcaption}
\usepackage{relsize}
\usepackage{caption}
\usepackage{multirow}
\usepackage{makecell}
\newcolumntype{L}[1]{>{\raggedright\let\newline\\\arraybackslash\hspace{0pt}}m{#1}}
\newcolumntype{C}[1]{>{\centering\let\newline\\\arraybackslash\hspace{0pt}}m{#1}}
\newcolumntype{R}[1]{>{\raggedleft\let\newline\\\arraybackslash\hspace{0pt}}m{#1}}

\wacvfinalcopy 

\def\wacvPaperID{****} 

\ifwacvfinal\fi
\begin{document}

\title{CNN-based Semantic Segmentation using Level Set Loss}

\author{
Youngeun Kim \thanks{These two authors contributed equally}\\
KAIST\\
\and
Seunghyeon Kim \footnotemark[1]\\
KAIST\\
\and
Taekyung Kim\\
KAIST\\
\and
Changick Kim\\
KAIST\\
\and
{\tt\small \{youngeunkim, seunghyeonkim, tkkim93, changick\}@kaist.ac.kr}
}

\maketitle
\ifwacvfinal\thispagestyle{empty}\fi

\begin{abstract}
Thesedays, Convolutional Neural Networks are widely used in semantic segmentation.
However, since {CNN-based segmentation networks} produce low-resolution {outputs} with rich semantic information, it is inevitable that spatial details (e.g., small objects and fine boundary information) of segmentation results will be lost.
%
To address this problem, motivated by a variational approach to image segmentation (i.e., level set theory), we propose a novel loss function called the level set loss which is designed to refine spatial details of segmentation results. 
To deal with multiple classes in an image, we first decompose the ground truth into binary images.
 Note that each binary image consists of background and regions belonging to a class.
Then we convert level set functions into class probability maps and calculate the energy for {each} class.
The network is trained to minimize the {weighted sum of the level set loss and the cross-entropy loss.} 
{The proposed level set loss improves the spatial details of segmentation results in a time and memory efficient way.}
Furthermore, our experimental results show that the proposed loss function achieves better performance than previous approaches.
\end{abstract}

\section{Introduction}
Semantic segmentation that allocates a semantic label to each pixel in an image  is one of the most challenging  tasks in computer vision.
However, traditional image segmentation methods \cite{threshold10b,  water, Levelset} are hard to address the task since they segment objects without semantic information.
Convolutional Neural Networks (CNNs) \cite{Lenet,Alexnet,CNN}  provide  a breakthrough for semantic segmentation  task.
Fully Convolutional Networks (FCNs) \cite{FCN,Deeplab} based on the CNN architecture is widely used thanks to its outstanding performance on semantic segmentation. 
However, as mentioned in \cite{Deeplabv3}, there are two challenges in CNN-based semantic segmentation networks: (1) consecutive pooling or striding causes the reduction of the feature resolution; (2) the networks are not aware of small objects.
Using dense CRFs \cite{CRF} as a post-processing step or modifying the network architecture with additional modules  \cite{FCN,PSPNet,Scalenet} are common solutions to these problems, but these approaches can be time-consuming and memory intensive \cite{Learning}.

\begin{figure}[t] 
	\centering
  	 	\begin{subfigure}[b]{0.11\textwidth}		
		\includegraphics[width=\textwidth]{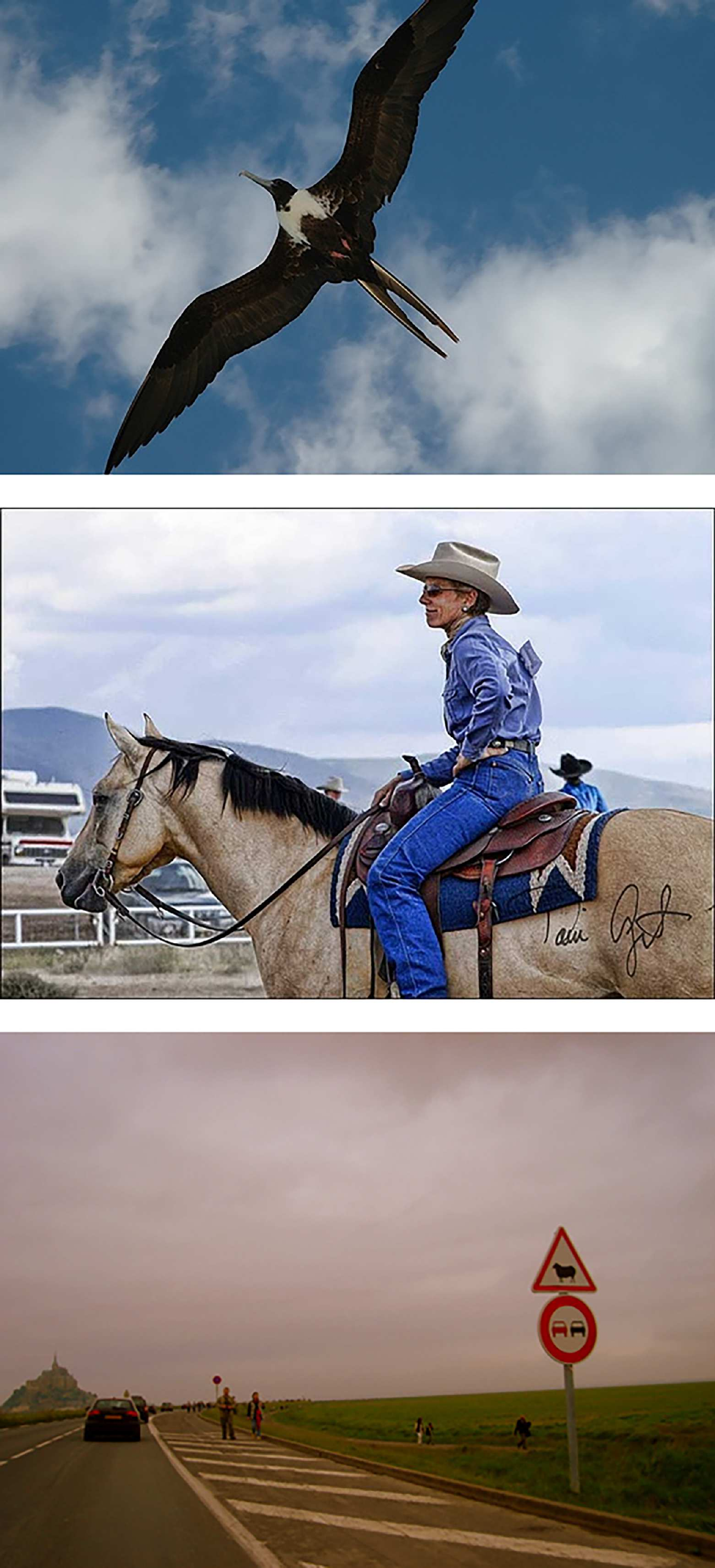}
		\caption{}
		\label{fig:quality_a}
	    \end{subfigure}
		\begin{subfigure}[b]{0.11\textwidth}		
		\includegraphics[width=\textwidth]{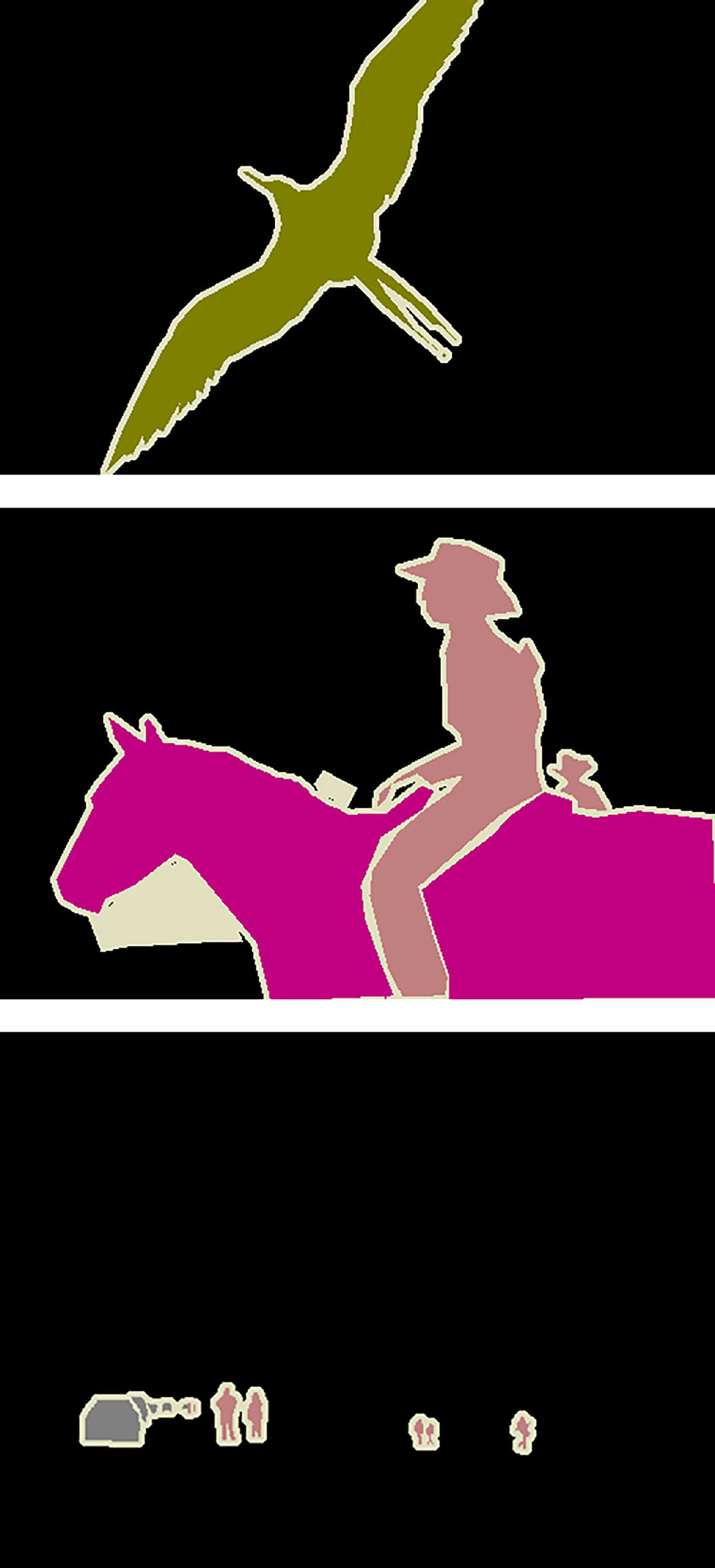}
		\caption{}
		\label{fig:quality_b}
	    \end{subfigure}
		\begin{subfigure}[b]{0.11\textwidth}		
		\includegraphics[width=\textwidth]{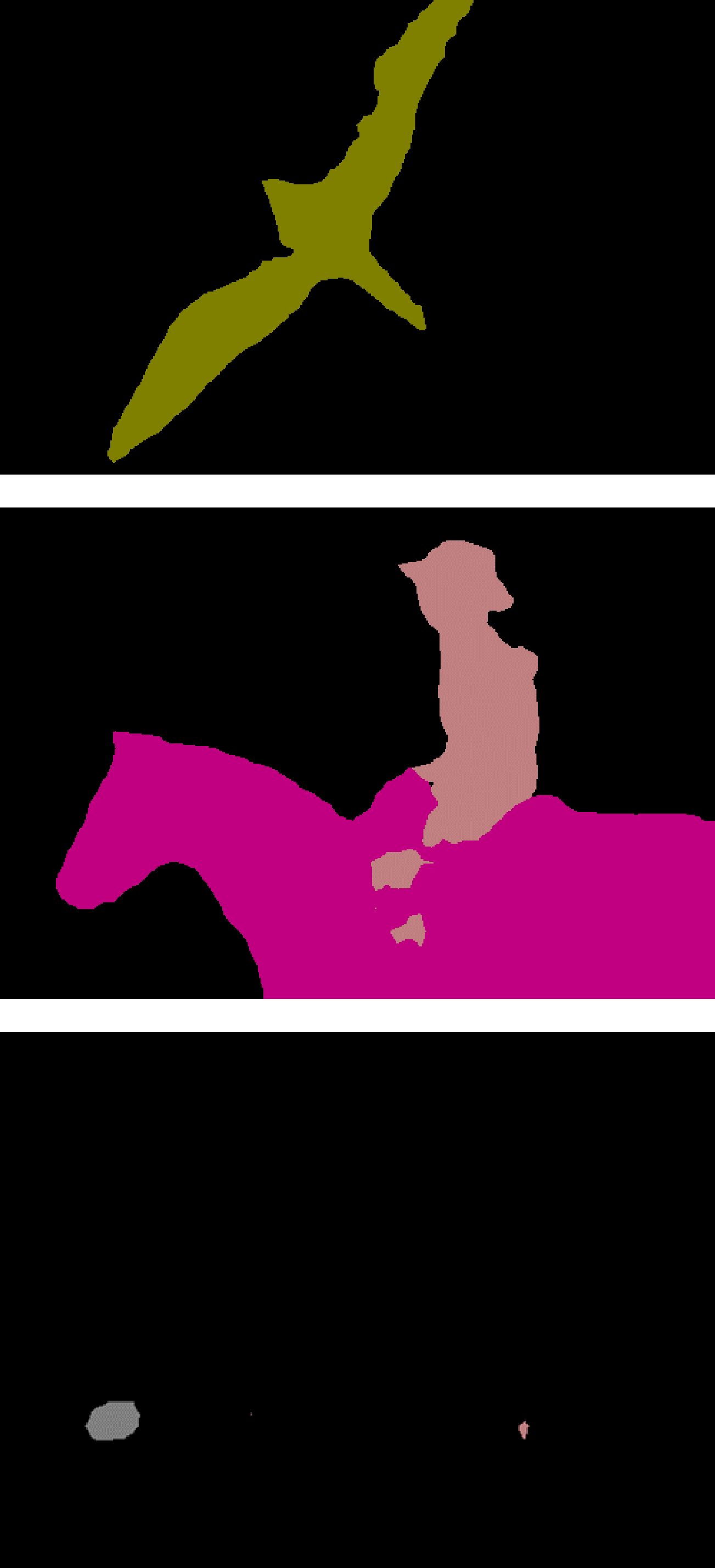}
		\caption{}
		\label{fig:quality_c}
	    \end{subfigure}
		\begin{subfigure}[b]{0.11\textwidth}		
		\includegraphics[width=\textwidth]{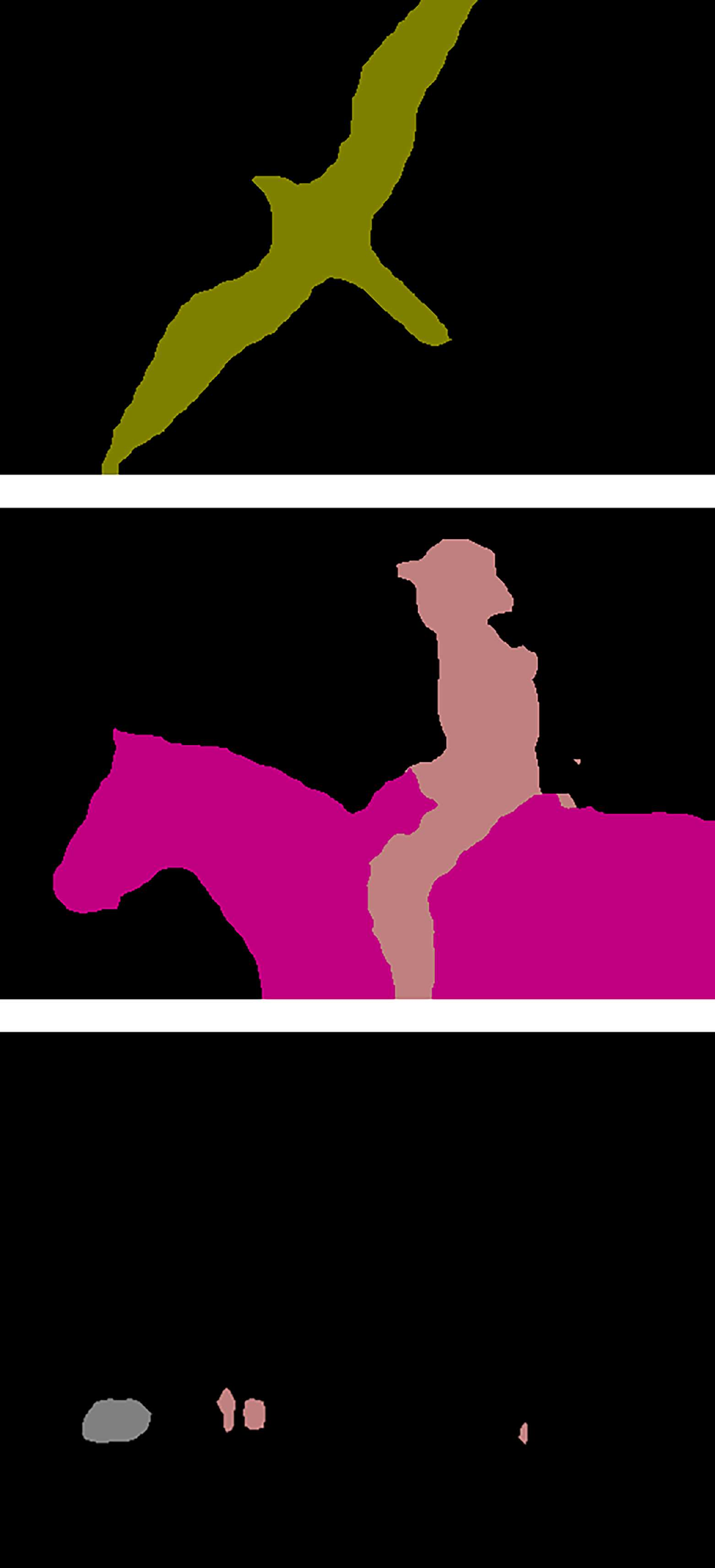}
		\caption{}
		\label{fig:quality_d}
	    \end{subfigure}
	   
	\caption{(a) Image. (b) Ground Truth. (c) DeepLab \cite{Deeplab}. \hspace{1ex}
	(d) Level set loss (Ours). Our loss alleviates the problems of semantic segmentation networks. First, proposed loss refines the boundaries of objects and fill missing parts of segmentation results (top and middle rows). Second, the level set loss encourages the network to segment small objects (bottom row).}
	\label{fig:Performance}
\end{figure}

{To overcome the aforementioned problems, in this paper, we introduce a loss function that utilizes spatial correlation in ground truth.
Until recently, most of the semantic segmentation frameworks use the cross-entropy loss.
However, the cross-entropy function calculates the loss at each pixel independently.
This is not desirable since segmentation network outputs are dense probability maps that contain semantic relation among pixels.}

We adopt the level set theory \cite{Levelset} to consider spatial correlation information of ground truth.
However, since the conventional level set function only separates the foreground and background of an image (i.e., single-class level set), it is hard to apply level set in a multi-class image.
To address this limitation, we separate the ground truth into the binary images of each class. 
We also note that the network outputs consist of the class probability maps.
For each class, by defining the probability map as a level set function for the binary image (i.e., class ground truth image), we can divide the multi-class level set problem into a number of single-class level set problems.
We exploit the level set function as the loss function in the training process. 
As shown in Fig. \ref{fig:Performance}, segmentation results with more sophisticated boundaries can be achieved by minimizing our level set loss.

Our contributions can be summarized as follows: 
i) We integrate the traditional level set method with a deep-learning architecture. 
To the best of our knowledge, ours is the first method that applies the level set theory on a CNN-based segmentation network.
ii) Compared to conventional approaches, the proposed end-to-end training method alleviates the additional effort (e.g.,   post-processing or network development) for preventing the reduction of spatial details.
iii) The proposed level set loss achieves better performance than previous approaches that suggest the loss function for semantic segmentation networks.
iv) To ensure the generality and efficiency of our level set loss, we apply the loss to two typical CNN-based architectures, FCN \cite{FCN} and  DeepLab \cite{Deeplab}.
 Experimental results show that the level set loss helps the segmentation network to learn spatial information of the objects.
 
The paper is organized as follows.
Section 2 presents the related work.
Section 3 describes the details of the level set loss. 
Section 4 presents the experimental results on widely used semantic segmentation networks. 
At the end of this paper, we conclude with our results and point out the future work.

\section{Related Work}

\subsection{Level Set Methods for Image Segmentation}

Active contour models, also known as snakes \cite{Snakes}, evolve a contour to detect objects in a given image using partial differential equations.
However, parametric snake models are sensitive to noise and initial contour location.
Among various approaches to solve this problem, level set methods are popular and widely used.
The main idea of level set methods is to find the level set function that minimizes the energy function.
As a result, the object boundary can be obtained by the zero level set.
Previous methods \cite{osher_levelset,narrow_band} calculate the energy function based on the edge information, which is usually sensitive to noise. 
To address this problem, a region-based level set method \cite{Levelset}  proposes the energy function that calculates the sum of pixel-intensity variance at the contour inside and outside.
The level set function is updated by iterative minimization of the energy function.
Finally, the zero level set represents the object boundary.
Since the region-based level set method is robust to image conditions such as noise or initial contour, this approach shows superior segmentation results than the prior methods.
\begin{figure*} [t]
	\centering
		\includegraphics[width=1\textwidth]{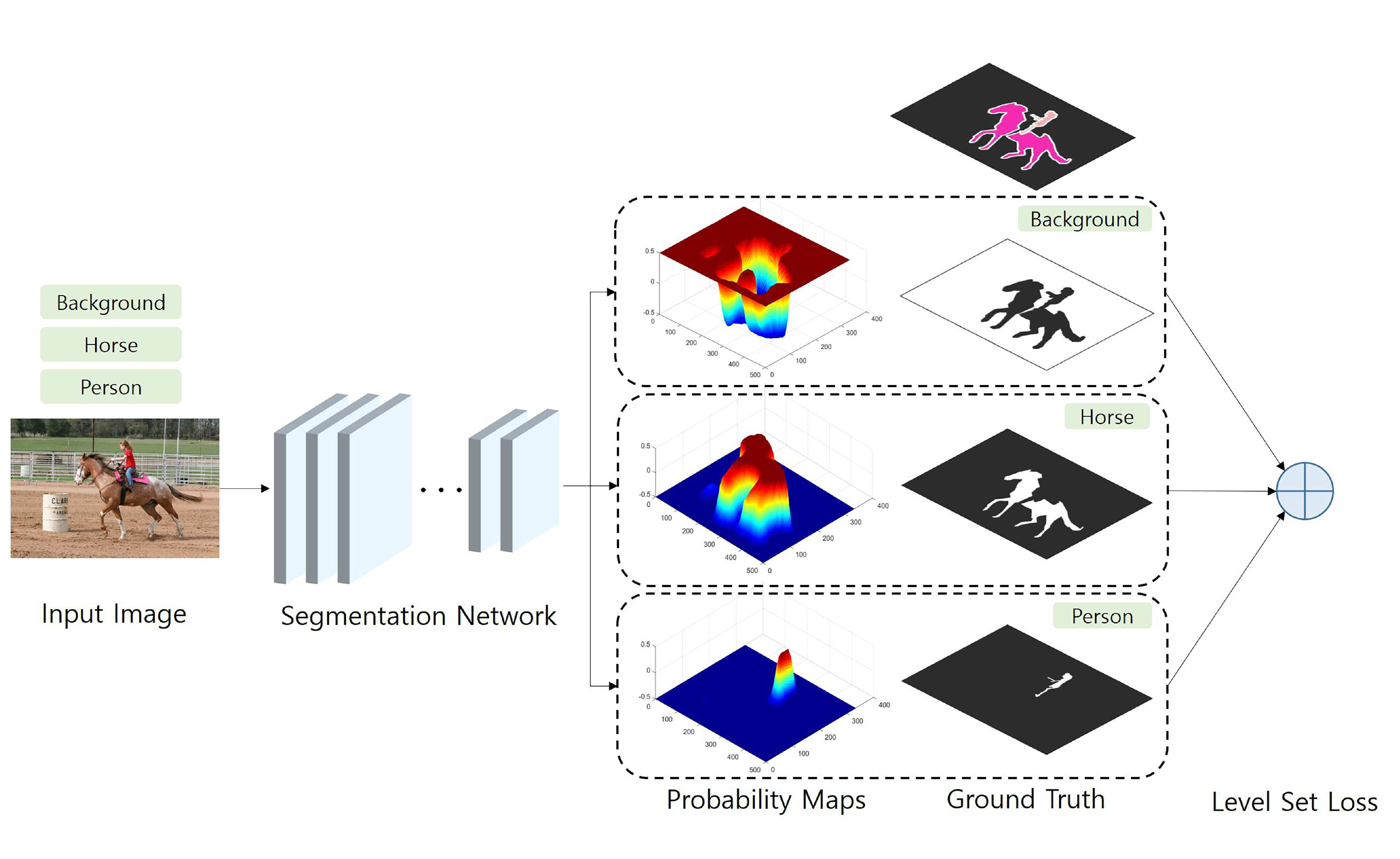}
	\caption{The training scheme of our level set loss. Our loss can be applied in an arbitrary CNN-based segmentation network.  The output probability maps are shifted into [-0.5, 0.5] so that it works as the level set function  $\phi$. We also decompose the ground truth into binary maps of each class (white is 1 and black is 0). For each class, we calculate the level set energy. We treat the sum of energy as  the loss function for training the segmentation network.}
	 \label{Loss Architecture}
\end{figure*}

\subsection{Semantic Segmentation}

Thanks to advances of  convolutional neural networks (CNNs), recent works achieve highly accurate results even in complex images \cite{VOC,COCO}.
However, there are two major problems with deep learning based segmentation approaches.
One is the reduction of  feature resolution due to consecutive pooling layers.
Another is the object size variability in realistic and complex images \cite{Deeplabv3}.

Recent works propose ways to deal with these problems.
FCN \cite{FCN} is aimed to address the problems by using encoder-decoder structure.
The decoder part recovers the object details and spatial information.
U-Net \cite{UNet} attaches skip connections between the features of encoder and decoder.
SegNet \cite{SegNet} stores pooling indices and reuses them in the decoder part.
DeepLab-v2 \cite{Deeplab} proposes atrous spatial pyramid pooling that adds multi-scale information from a parallel structure.
PSPNet \cite{PSPNet}  improves the performance via pyramid pooling module, which extracts global context information by aggregating different region based contexts.
 Deeplab-v3+ \cite{Deeplabv3+} attaches decoder part after atrous spatial pyramid pooling module.
 The encoder part with atrous convolution extracts rich semantic features and the decoder part recovers the object boundary.

With the development of networks, several techniques are also proposed to handle the challenges in semantic segmentation.
Pre-trained weight initialization is pretty important to produce a satisfying performance. 
This technique gives the network prior knowledge of object classification.
Most semantic segmentation networks use pre-trained weight on ImageNet \cite{ImageNet}, and some of them also use COCO \cite{COCO} and JFT \cite{JFT} for a higher score. 
Data augmentation is also widely used to increase the amount of data and to avoid local minima. 
These techniques are more critical for semantic segmentation due to lack of data. 
Conditional Random Fields (CRFs) \cite{Deeplab,CRF}  are the common post-processing method  for boundary refinement. 
This method improves performance especially along the boundaries of the objects, but is sensitive to their parameters.
Also, multi-scale inputs (MSC) \cite{Scalenet} provide performance increase.
For multi-scale processing, inputs are given to the network at scale $ = \lbrace 0.5, 0.75, 1 \rbrace$ and the output is determined by selecting the maximum output value across scales for each pixel.
 
\subsection{Level Set with Deep Learning Framework}

Few works adopt the idea of the level set into the deep learning framework.
In other fields, \cite{hu2017deep} proposes a deep level set network for saliency object detection.
They use the level set theory to refine the saliency maps.
Furthermore, they apply super-pixel filtering to help the refinement.
Since their method works only with saliency maps, the approach is hard to apply in multi-class and multi-object images.

For semantic segmentation, \cite{CRLS} proposes Contextual Recurrent Level Set (CRLS).
In this work, the curve evolution is presented in a time series, and the level set method is reformulated as Recurrent Neural Network (RNN).
Since the level set method is hardly applied in multi-class images, they utilize an object detection network to obtain single object images.
While this method needs an auxiliary network, our approach can improve the performance without any additional networks or architectural changes.

\subsection{Loss Functions for CNNs}
Deep learning framework usually uses the cross-entropy loss for a cost function. The cross-entropy loss gives satisfying results on various tasks (e.g., classification, object detection, semantic segmentation). However, there are some missing points with the cross-entropy loss when it is used for object detection or semantic segmentation. Recently, there have been  several attempts to increase the performance of scene understanding  by adding loss terms or changing the structure of loss layers.

In object detection, \cite{Focal} suggests Focal Loss to solve the extreme foreground-background class imbalance problem in one-stage object detection.
Their Focal Loss adopts  the cross-entropy loss to focus on hard negative examples.
With RetinaNet proposed in  \cite{Focal}, their loss shows the improvement in object detection area.
For semantic segmentation task, \cite{LMP} also  tries to deal with the class imbalance problem.
By applying a max-pooling concept to the loss,  they re-initialize the weights of the individual pixel based on the value of loss functions.
\cite{LAD}  addresses the problem of  the pixel-wise loss  as we suggest in this paper.
By using the locally adaptive learning estimator, they enforce the network to learn the inter and intra class discrimination.

\section{Our Method}
In this section, we introduce our novel loss for CNN-based semantic segmentation networks.
We first review the classic level set method \cite{Levelset}, which is important to understand our work.
Then, our proposed loss is described in details.

\subsection{Region-based Level Set Methods}
The authors of \cite{Levelset} propose the region-based level set method for image segmentation.
The method minimizes the energy function which is defined by
\begin{equation} \label{energy_function}
\begin{aligned}
	F(c_1, c_2, \phi) 	& = \mu\cdot Length(\phi) + \nu\cdot Area(\phi) \\
	& + \lambda_1\int_{\Omega} {\vert u_0(x,\,y)-c_1\vert}^2 \, H(\phi(x,\,y))\, dx\, dy \\
	& + \lambda_2\int_{\Omega} {\vert u_0(x,\,y)-c_2\vert}^2 \, (1-H(\phi(x,\,y)))\, dx\, dy,
	\end{aligned}
\end{equation}
where $\mu\geq0, \nu\geq0, \lambda_1, \lambda_2>0$ are fixed parameters, $\Omega$ is the entire domain of the given image, $u_0(x,y)$ is the pixel value at location $(x,y) \in \Omega$, and $\phi$ is the level set function. $Length(\phi)$ and $Area(\phi)$ are regularization terms with respect to the length and the inside area of the contour. 
$H$ is the Heaviside Function (HF), 
\begin{equation}
	H(z)=\begin{cases}
		1, & \text{$z \geq 0$} \\
		0, & \text{$z < 0$}.
	\end{cases}
	\label{HF}
\end{equation}
$c_1$ and $c_2$ are constant functions of $\phi$ that indicate the mean pixel value of interior and exterior of the contour respectively.
\begin{equation}
    \label{c}
    \begin{gathered}
        c_1(\phi) = \frac{\int_{\Omega}u_0(x,\,y) \, H(\phi(x,\,y)) \, dx \, dy}{\int_{\Omega}H(\phi(x,\,y)) \, dx \, dy},
        \\
        c_2(\phi) = \frac{\int_{\Omega}u_0(x,\,y) \, (1-H(\phi(x,\,y))) \, dx \, dy}{\int_{\Omega}(1-H(\phi(x,\,y))) \, dx \, dy}.
    \end{gathered}
\end{equation}

\subsection{Deep Level Set Loss}

We utilize the energy function of \cite{Levelset} for the semantic segmentation task with deep learning framework. 
{Since the classic level set method is limited in representing semantic information}, we decompose multi-class semantic segmentation into several single-class segmentation by reconstructing the dense binary ground truth for each object.
For a given input image, let $G$ be the given segmentation ground truth and $L$ be the set of classes that exist in the image. We generate the reconstructed ground truth $G_l$ for class $l \in L$ by remaining object region of concern in $G$ as foreground and replacing the others with background. Note that we also generate binary dense ground truth $G_0$ for background class in $L$. \\

\begin{figure}[t]
	\centering
	\begin{subfigure}[b]{0.2\textwidth}
		\includegraphics[width=\textwidth]{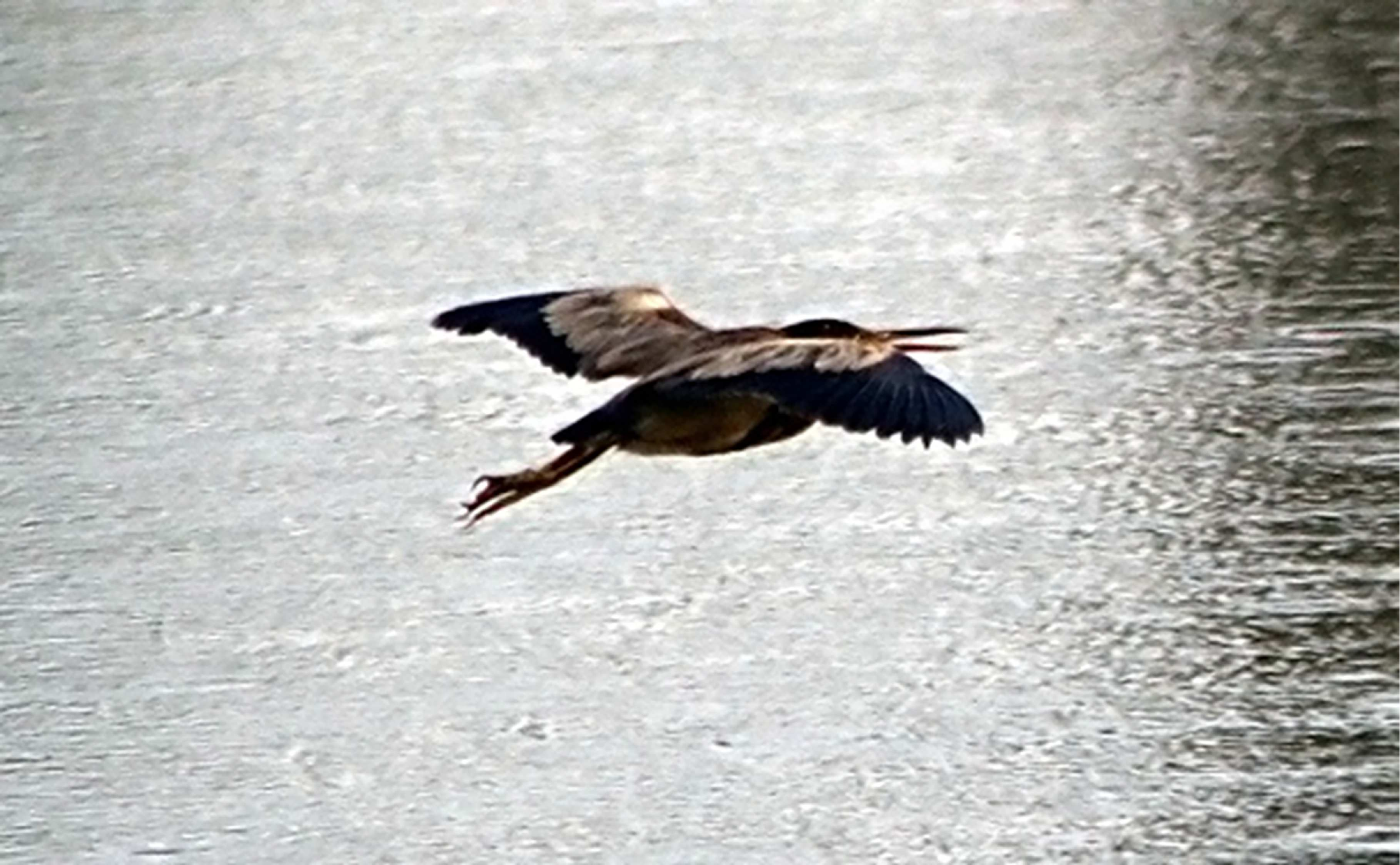}
		\caption{Image}
		\label{fig:bird_input}
	\end{subfigure}
	\begin{subfigure}[b]{0.2\textwidth}
		\includegraphics[width=\textwidth]{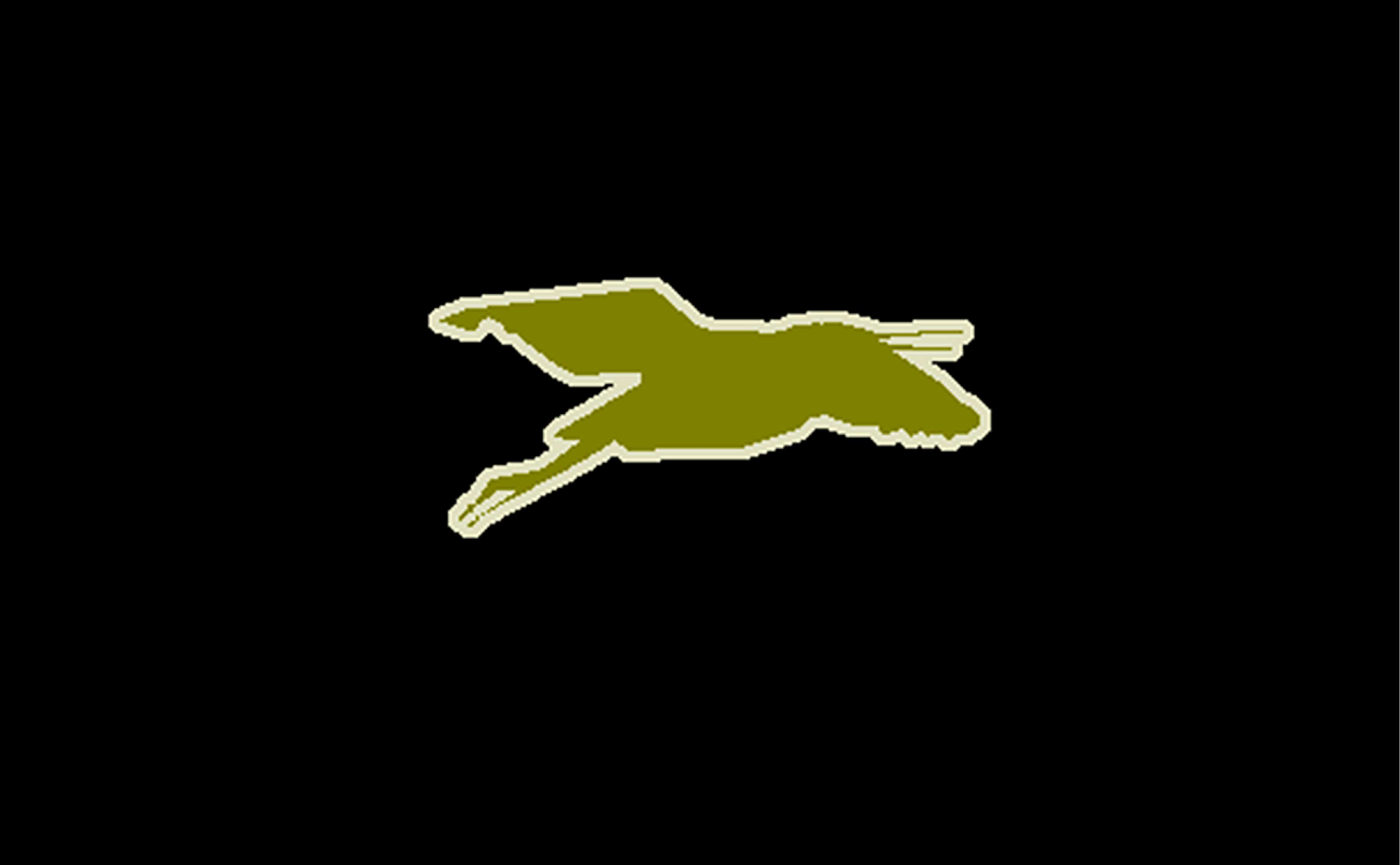}
		\caption{Ground Truth}
		\label{fig:bird_gt}
	\end{subfigure}
	\begin{subfigure}[b]{0.2\textwidth}
		\includegraphics[width=\textwidth]{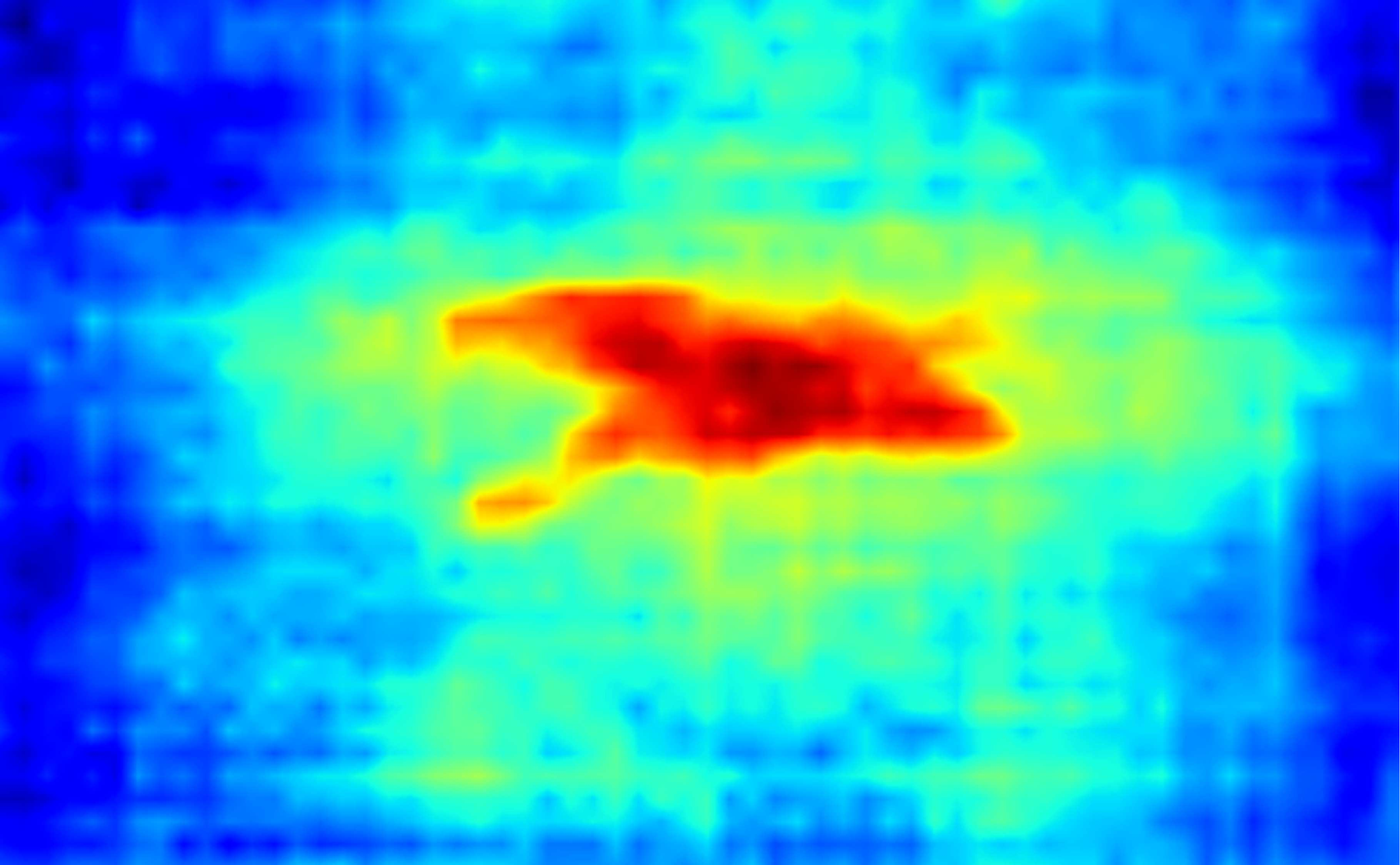}
		\caption{CE loss only}
		\label{fig:bird_CE}
	\end{subfigure}
	\begin{subfigure}[b]{0.2\textwidth}
		\includegraphics[width=\textwidth]{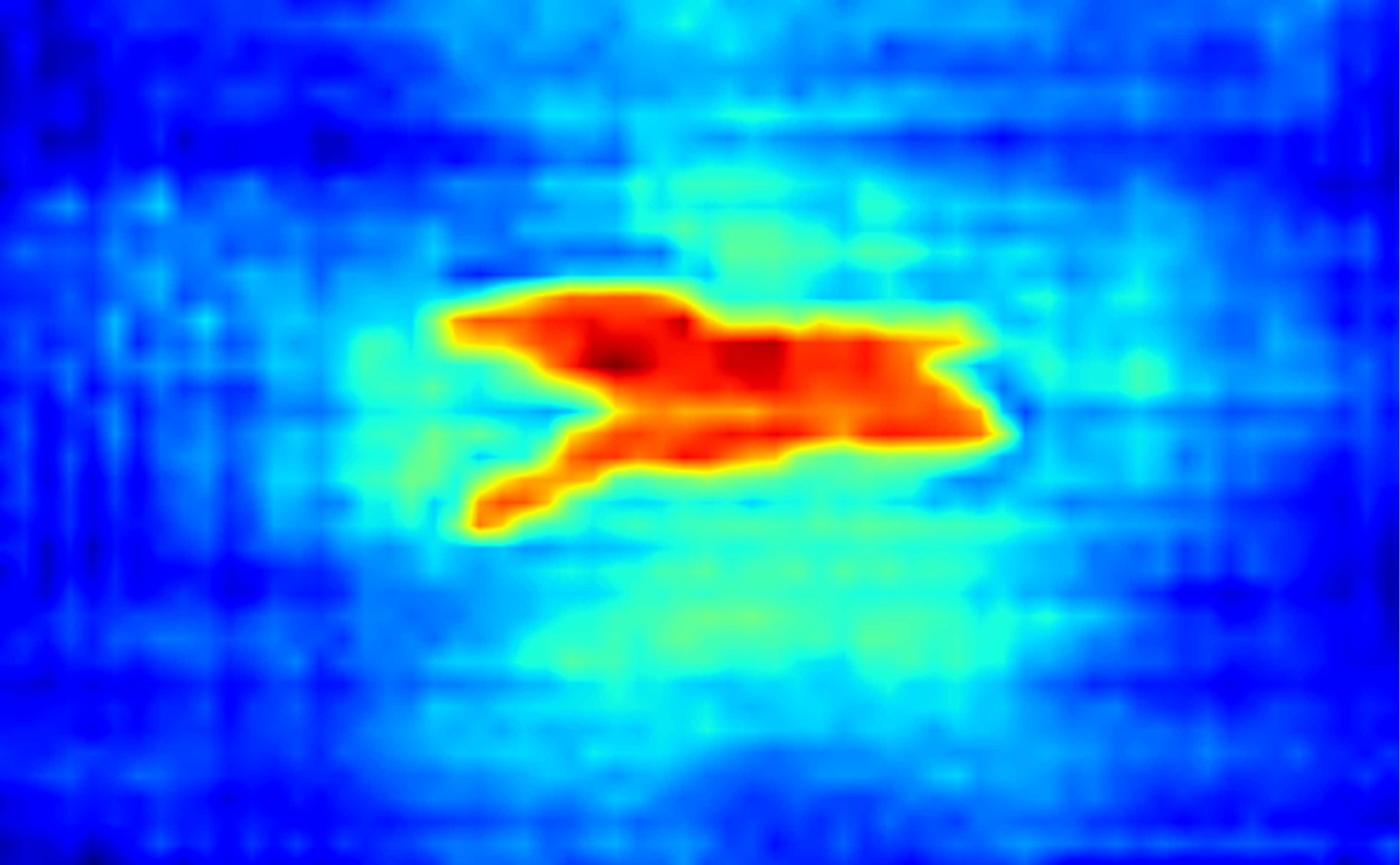}
		\caption{with LS loss}
		\label{fig:bird_LS}
	\end{subfigure}
	\caption{Visualization of the level set function (i.e., probability map). We show  the probability map of class Bird (blue is 0 and red is 1). Compared to (c), the network trained with level set loss (d) provides clearer boundary. Best view in color.}
	\label{fig:bird}
\end{figure}

To apply the classic level set method for deep learning, we first set the parameters in eq. \eqref{energy_function} as done in \cite{Levelset}, $\nu=0,\,\lambda_1=\lambda_2=1$.
Also, as mentioned in \cite{Levelset}, the $Length(\phi)$ term is sensitive to the size of the object. 
Since the input images have multiple size objects, we set $\mu$ as zero.
Our proposed level set loss is formulated as follows:
\vspace{4ex}
\begin{small}
\begin{equation} \label{ls_loss}
\begin{aligned}
E_{LS}(\phi,\,G) & = \mathlarger{\mathlarger{\sum}}_{l \in L} \Bigg{(} \,
\int_{\Omega_l} {\vert{G_l(x,\,y)-c_{l,1}}\vert}^2 H^*_{\epsilon}({\phi_l}(x,\,y))\,dx\,dy \\
 & + \, \int_{\Omega_l} {\vert{G_l(x,\,y)-c_{l,2}}\vert}^2 (1-H^*_{\epsilon}({\phi_l}(x,\,y)))\,dx\,dy \Bigg{)},
\end{aligned}
\end{equation}
\end{small}
where $\Omega_l$ is an entire domain of $G_l$, and the level set function $\phi$ is a shifted dense probability map that is estimated from the  segmentation network e.g., ${\phi_l}(x,\,y)={P_l}(x,\,y)-0.5 \in [-0.5,\, 0.5] $ with the output probability map $P_l$ for class $l$.
Note that we apply the energy function on the reconstructed dense ground truth for each existing class instead of the input image.
Since the objects in an image {may have} high color variance, it is not desirable to apply the level set function.
So, we replace $u_0$ to $G_l$ for a reliable training process.
\(c_{l,1}\) and \(c_{l,2}\) represent  average intensity of binary ground truth map $G_l$ for contour inside and outside.

\begin{equation}
    \label{modify_c}
    \begin{gathered}
        c_{l,1}(\phi) = \frac{\int_{\Omega}G_l(x,\,y) \, H^*_{\epsilon}(\phi_l(x,\,y)) \, dx \, dy}{\int_{\Omega}H^*_{\epsilon}(\phi_l(x,\,y)) \, dx \, dy},
        \\
        c_{l,2}(\phi) = \frac{\int_{\Omega}G_l(x,\,y) \, (1-H^*_{\epsilon}(\phi_l(x,\,y))) \, dx \, dy}{\int_{\Omega}(1-H^*_{\epsilon}(\phi_l(x,\,y))) \, dx \, dy}.
    \end{gathered}
\end{equation}

\begin{figure*}[t]
	\centering
	\begin{subfigure}[b]{0.32\textwidth}		
		\includegraphics[width=\textwidth]{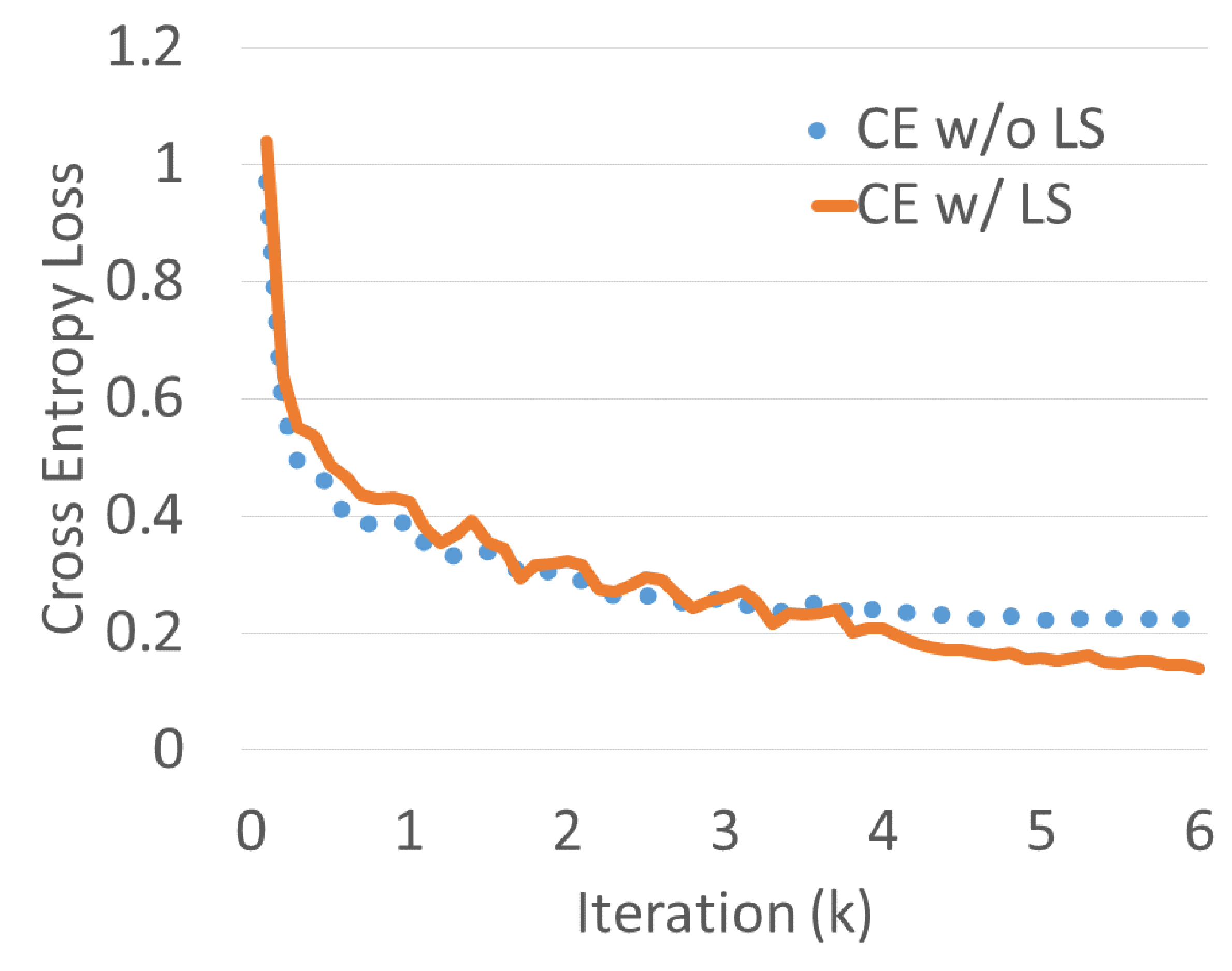}
		\caption{CE loss}
		\label{fig:CELoss}
	\end{subfigure}
	\begin{subfigure}[b]{0.32\textwidth}	
		\includegraphics[width=\textwidth]{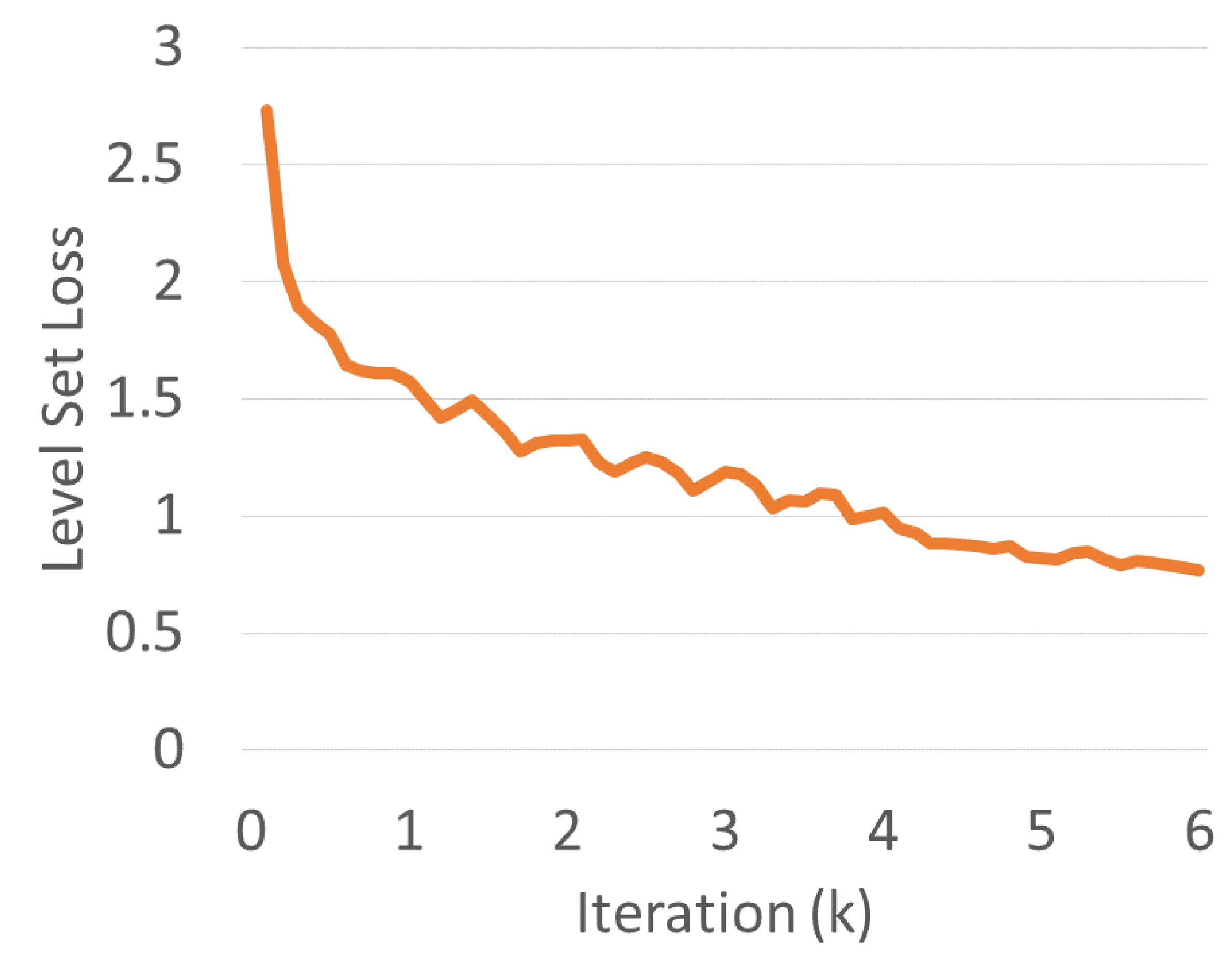}
		\caption{LS loss}
		\label{fig:mIoU}
	\end{subfigure}
	\begin{subfigure}[b]{0.32\textwidth}
		\includegraphics[width=\textwidth]{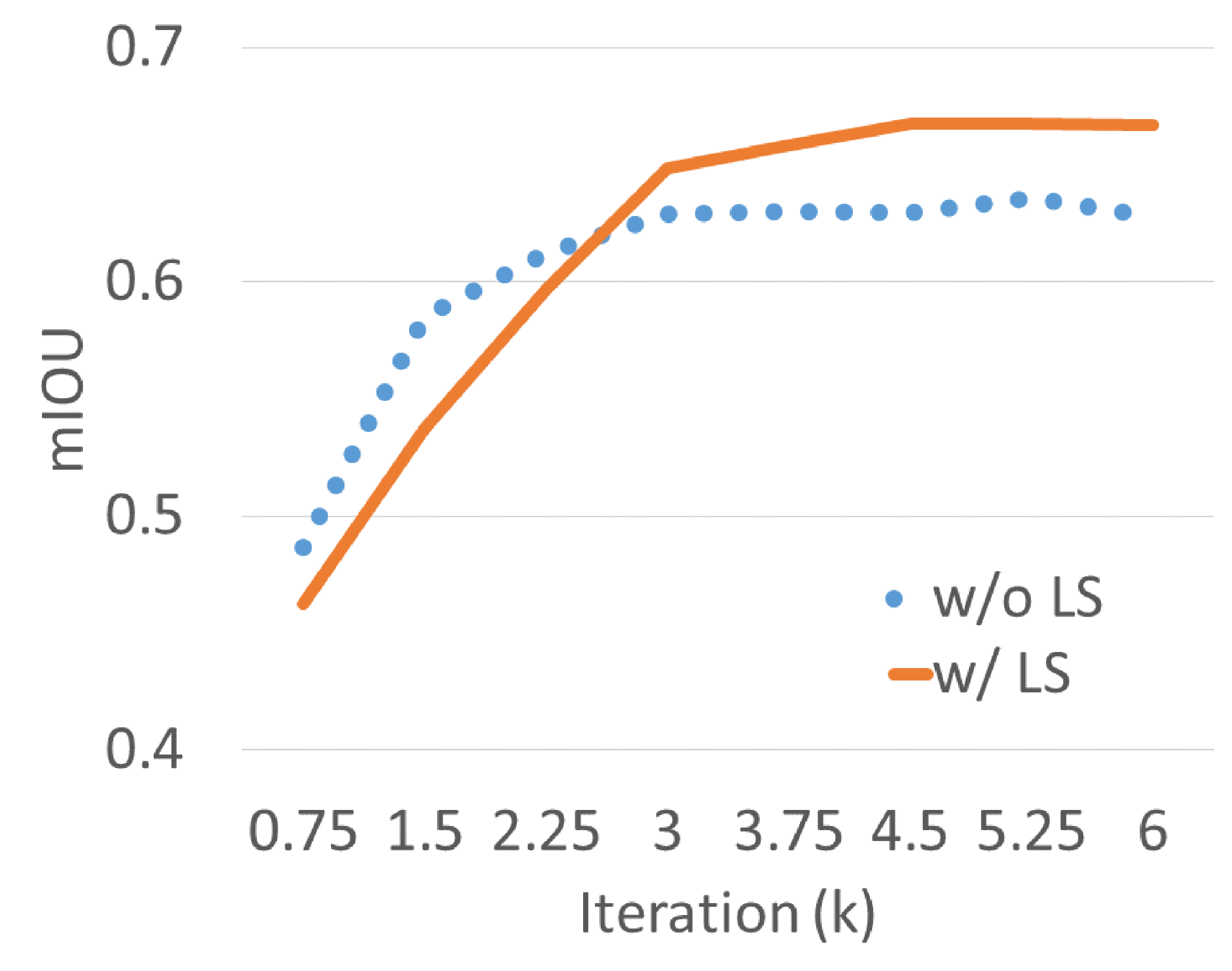}
		\caption{mIOU}
		\label{fig:LSloss}
	\end{subfigure}
	\caption{Three graphs of two losses and mIOU over training iteration. (a) As expected, the cross-entropy loss is decreased. (b) LS loss is also decreased with increasing the mIOU score. (c) A model with higher mIOU has lower LS loss. }
		\label{Statistics}
\end{figure*}

In our method, we propose the Modified Approximated Heaviside Function (MAHF)  $H^*$.
We note that using $tanh$ as an activation function shows high performance for deep learning architectures \cite{Efficient,DCGAN}.
Therefore, contrary to original  AHF proposed in \cite{Levelset}, we adopt $tanh$  to achieve the similar effect as using Heaviside Function, eq. \eqref{HF}.
As expected, we observe that employing $tanh$ performs better than using HF and original AHF.
\begin{equation} \label{H_star}
        H^*_{\epsilon}(z) = \frac{1}{2}(1 + tanh(\frac{z}{\epsilon})).
\end{equation}

Figure  \ref{Loss Architecture} shows the whole training scheme for calculating the level set loss.
The network can be end-to-end trained using backpropagation with our proposed loss. 
Since our loss function is differentiable with respect to network outputs, as shown in eq. \eqref{ls_loss_diff},  gradients are available for all training processes.
\begin{equation} \label{ls_loss_diff}
    \begin{gathered}
        \frac{\partial E_{LS}}{\partial \phi_l} = \delta_{\epsilon}^*(\phi_l)\Bigg{[}(G_l-c_{l,1})^2-(G_l-c_{l,2})^2\Bigg{]}, \\
        \delta^*_{\epsilon}(z) = \frac{\partial H^*_{\epsilon}(z)}{\partial z} = \frac{1}{2\epsilon}(1 - tanh(\frac{z}{\epsilon}))(1 + tanh(\frac{z}{\epsilon})).
    \end{gathered}
\end{equation}

To help  the network to learn more discriminative features of objects, we combine our loss function with the cross-entropy loss as follows:
\begin{equation} \label{total_loss}
	Loss = E_{CE}(P,\,G) + \lambda \cdot E_{LS}(\phi,\,G),
\end{equation}
where $\lambda$ is a parameter for weighting the level set loss. 
With eq. \eqref{total_loss}, the level set loss decreases as mIOU increases,  as shown in Fig. \ref{Statistics} (b) and (c).
Also,  CE loss with LS loss reaches lower value than that of baseline (without LS loss), as shown in Fig. \ref{Statistics} (a).
These statistics imply that the level set loss trains the network in a different way from CE loss.
This is possible due to the property of  the energy function,  eq. \eqref{energy_function}, which considers overall spatial information of an image.

\section{Experimental Results}

In this section, we show experimental results of proposed level set loss.
We compared our method with different approaches (e.g., encoder- decoder structure network and CRFs).
We selected FCN \cite{FCN} as the representative architecture for experimenting on the encoder-decoder structure network and DeepLab \cite{Deeplab} as the representative architecture for experimenting on the network with CRFs as post-processing.
%
%
The comparison with  previous similar approaches (i.e., proposing  loss function for semantic segmentation) are shown in Section 4.6.
Our proposed loss achieves better segmentation results than others.
 We present  {analyses of the level set loss} on hyperparameters and MAHF in Section 4.7.

\subsection{Evaluation Metric and Dataset}

The performance was measured in terms of pixel intersection-over-union (IOU) averaged across the every classes.
We carry out experiments on three semantic segmentation public datasets (PASCAL VOC 2012, PASCAL-Context, Cityscapes).

\subsubsection{PASCAL VOC 2012}
We evaluated our framework on the PASCAL VOC 2012 dataset \cite{VOC}, which are comprised of 20 foreground object classes and one background class.
Like previous works, we used extra annotations \cite{augmentation} so that the dataset contains 10,582 training, 1,499 validation and 1,456 test images.
The experimental results are reported in Table \ref{result on FCNs}.

\subsubsection{PASCAL-Context}
The PASCAL-Context \cite{context} dataset is another popular dataset for semantic segmentation evaluation.
The dataset contains 59 object classes and one background class.
We use 4998 images for training and 5105 images for testing.
The results are shown in Table  \ref{context}.

\subsubsection{Cityscapes}
The Cityscapes dataset \cite{city} consists of street scene images
from 50 different cities.
The dataset contains pixel-level annotations of cars, roads, 
pedestrians, motorcycles, etc.
In total the dataset considers 19 classes.
The dataset provides 2975 images for training and the 500 images for validating.
The results are shown in Table \ref{City}.

\subsection{Implementation Details}
Our implementation is based on the Pytorch \cite{paszke2017automatic} and the Tensorflow \cite{Tensorflow}.
We trained the network by the standard SGD algorithm of  \cite{Alexnet}.
Also, ResNet-based network was pre-trained with MS-COCO \cite{COCO}.
For FCN, initial learning rate, batch and crop size were set to (0.004, 10, 512).
 The network was trained with the momentum 0.9 and the weight decay 0.0001. 
Random horizontal flip was used for data augmentation.
 As training DeepLab framework,  initial learning rate, batch and crop size were set to (0.004, 10, 512) and (0.00025, 10, 321) for DeepLab-LargeFOV and DeepLab-ResNet101, respectively.
 To train VGG16-based network (i.e., DeepLab-LargeFOV), we set the momentum 0.9, the weight decay 0.0001 and we applied horizontal flip to input images.
 Otherwise, DeepLab-ResNet101 was trained with the momentum 0.9, the weight decay 0.0005 and we applied random crop and horizontal flip to input images.
In addition,  we used the ``poly" learning rate policy where the current learning ratio equals initial learning rate multiplied by $(1-\frac{iter}{max\_iter})^{0.9}$ for training DeepLab-ResNet101.

\setlength{\tabcolsep}{4pt}
\begin{table} [t]
\begin{center}
\begin{tabular}{l|c|cc}
\Xhline{3\arrayrulewidth}
 Method               & Baseline & CRFs & LS loss \\ \hline
                      
FCN-32s-ResNet101     & 65.1     & -       &  68.9  \\
FCN-16s-ResNet101     & 68.4     & -       & 69.2       \\
FCN-8s-ResNet101      & 68.5     & -      &  69.3         \\ \hline \hline
DeepLab-LargeFOV      & 62.8     &  67.3  & 66.7       \\
DeepLab-MSC-LargeFOV      & 64.2     & 68.1   & 67.2       \\
DeepLab-ResNet101     & 75.1     &  76.3   &  76.5   \\
DeepLab-MSC-ResNet101     &   76.0   &  77.7  &   77.3\\ \Xhline{3\arrayrulewidth}

\end{tabular}
\caption{Performance comparison of our level set loss on representative architectures. We train the FCN-ResNet101 \cite{FCN} and DeepLab \cite{Deeplab}. 
The Pascal VOC 2012 validation set is used for evaluation.    }
\label{result on FCNs}
\end{center}
\end{table}
\setlength{\tabcolsep}{1.4pt}


\subsection{Running Time for Level Set Loss}
We report execution time on NVIDIA GTX 1080Ti and INTEL i7-8700 processor with 16G RAM.
To measure training time, we adopt DeepLab-largeFOV as a baseline network.
In training phase, it took 0.133 sec/image with the cross-entropy only.
With our level set loss, it took 0.157 sec/image.

\setlength{\tabcolsep}{6pt}
\begin{table}[]
\begin{center}
\begin{tabular}{l|c|cc}
\Xhline{3\arrayrulewidth}

Method            & Baseline & CRFs & LS loss \\ \hline
FCN-8s-ResNet101  & 41.0     & -       & 42.2       \\
DeepLab-LargeFOV  & 37.1     & 39.7    & 39.4       \\
DeepLab-ResNet101 & 44.7     & 45.7    & 45.5       \\ \Xhline{3\arrayrulewidth}

\end{tabular}
\caption{mIOU performance on the Pascal-Context test set.}
\label{context}
\end{center}
\end{table}


\begin{table}[]
\begin{center}
\begin{tabular}{l|c|cc}
\Xhline{3\arrayrulewidth}
Method            & Baseline & CRFs & LS Loss \\ \hline
FCN-8s-ResNet101  & 64.7     & -       & 65.5       \\
DeepLab-LargeFOV  & 62.9     & 64.1    & 64.0       \\
DeepLab-ResNet101 & 68.8     & 69.8    & 69.8       \\ \Xhline{3\arrayrulewidth}
\end{tabular}
\caption{mIOU performance on the Cityscapes validation set.}
\label{City}
\end{center}
\end{table}

\subsection{Effects of Level Set Loss}

\subsubsection{Fully Convolutional Networks (FCN)}
Table  \ref{result on FCNs} summarizes segmentation results (mIOU) of FCNs based on ResNet101 with and without LS loss. 
The baselines were trained only with the cross-entropy loss. 
Using the proposed LS loss consistently yielded performance gains over different FCNs.
Especially, FCN-32s-ResNet101 with LS loss achieved mIoU of 68.9\%, which outperforms the baseline by 3.8\%. 
Note that despite of the systematic low resolution, mIoU of FCN-32s-ResNet101 with LS loss exceeds that of FCN-8s-ResNet101 baseline by 0.4\%.
It shows  that the LS loss not only guides the object boundary but also encourages learning spatial correlation information.
Qualitative results in Fig. 5 further prove the spatial context awareness of the network trained by LS loss.

\subsubsection{DeepLab}
For DeepLab implementation, we used the same training scheme as \cite{Deeplab}.
To show generality of our loss function, we evaluated our loss on DeepLab-LargeFOV and DeepLab-ResNet101. \ref{result on FCNs}
As shown in Table \ref{result on FCNs}, using the LS loss on DeepLab-LargeFOV increases the performance about 4\%.
For DeepLab-ResNet101, LS loss achieves about 1\% mIOU improvement.
%
%
In our experimental results, employing the LS loss achieves comparable performance to using CRFs  in DeepLab frameworks.
As shown in Fig. \ref{qualitative_results}, the results of CRFs show more accurate object boundaries than the results of the LS loss.
However, LS loss supervises the network to detect complex or small objects.
Furthermore, our loss performs end-to-end training, while CRFs are used as a post-processing method which requires extra computation.

\setlength{\tabcolsep}{1.2pt}
\begin{table}[]
\begin{center}
\begin{tabular}{c|c|c|cc}
\Xhline{3\arrayrulewidth}
       DeepLab                         &  time : (sec/img)            & LS loss        & CRFs\scriptsize{-iter(5)} & CRFs\scriptsize{-iter(10)}                  \\ \hline \hline
\multirow{3}{*}{LargeFOV}  & time & 0.025 &  0.272 & 0.387 \\ \cline{2-5} 
& relative time & ($\times 1$)&   ($\times 10.8$)  &  ($\times 15.4$)                        \\ \cline{2-5} 
                                   & mIOU           & 66.7            & 66.5         & 67.3 \\ \hline
\multirow{3}{*}{ResNet101} & time & 0.054 &0.315 &0.452 \\ \cline{2-5} 
 & relative time & ($\times 1$)    &($\times 5.8$)          &($\times 8.4$)                      \\ \cline{2-5} 
                                   & mIOU           &76.5    & 76.1         & 76.3                           \\ \Xhline{3\arrayrulewidth}
\end{tabular}
\caption{Runtime comparison between using level set loss and using CRFs on the Pascal 2012 validation dataset.
“iter(\textit{k})” means that the number of iteration in CRFs. }
\label{Runtime}
\end{center}
\end{table}

\subsection{Comparison with CRFs}

Using CRFs gives similar performance to ours as shown in Table \ref{result on FCNs},     \ref{context}, \ref{City}.
However, using CRFs as post-processing requires extensive computational time.
On the other hand, our algorithm is end-to-end approach and does not have additional computation at inference time.
We experimentally compare the time consumption between using CRFs and the level set loss.
In Table \ref{Runtime}, the results show that our method achieves a speed increase, which is higher than 5$\times$ compared to CRFs.
Furthermore, the performance of our method is better than using CRFs with less iteration.

\subsection{Comparison with Other Loss Functions}

There are only a few works that propose the loss function for  semantic segmentation task.
For a fair comparison with previous loss functions,  we presented the results on DeepLab-ResNet101.
Here, we do not use both multi-scale input (MSC)  and CRFs, as done in  \cite{LMP} and  \cite{LAD}.
We also used the Pascal VOC 2012 segmentation validation dataset for evaluation.

As shown in Table \ref{result_loss}, the network trained with our level set loss is better than previous methods.
Our loss function (we use $\epsilon=1/20$,  $\lambda = 4\times 10^{-4}$ for hyper parameters)   shows much improved mIOU that is  1.4\% higher than the  baseline.
The closest competing method is LMP \cite{LMP}, which achieves 1.2\%  higher than the baseline.
By training the network with spatial information of ground truth, our level set loss significantly boosts the mIOU.

\setlength{\tabcolsep}{4pt}
\begin{table} [t]
\begin{center}
\begin{tabular}{l|c}
\Xhline{3\arrayrulewidth}
                       Method & mIOU \\ \hline
                      
DeepLab-ResNet101 (Baseline)       & 75.1  \\
 \(L^{AD}\)   \cite{LAD}						& 76.1    \\ 
 LMP \cite{LMP}     		   						& 76.3        \\ \hline 
 Level Set Loss (Ours)    						&{\bf 76.5 }\\  \Xhline{3\arrayrulewidth}

\end{tabular}
\caption{Performance comparison with other loss functions for semantic segmentation networks. We present the mIOU reported in \cite{LMP} and \cite{LAD}.
The Pascal VOC 2012 validation set is used for evaluation.}
\label{result_loss}
\end{center}
\end{table}
\setlength{\tabcolsep}{1.4pt}


\subsection{Analysis of the MAHF }
\begin{figure} [t] 
	\centering
		\includegraphics[width=0.41 \textwidth]{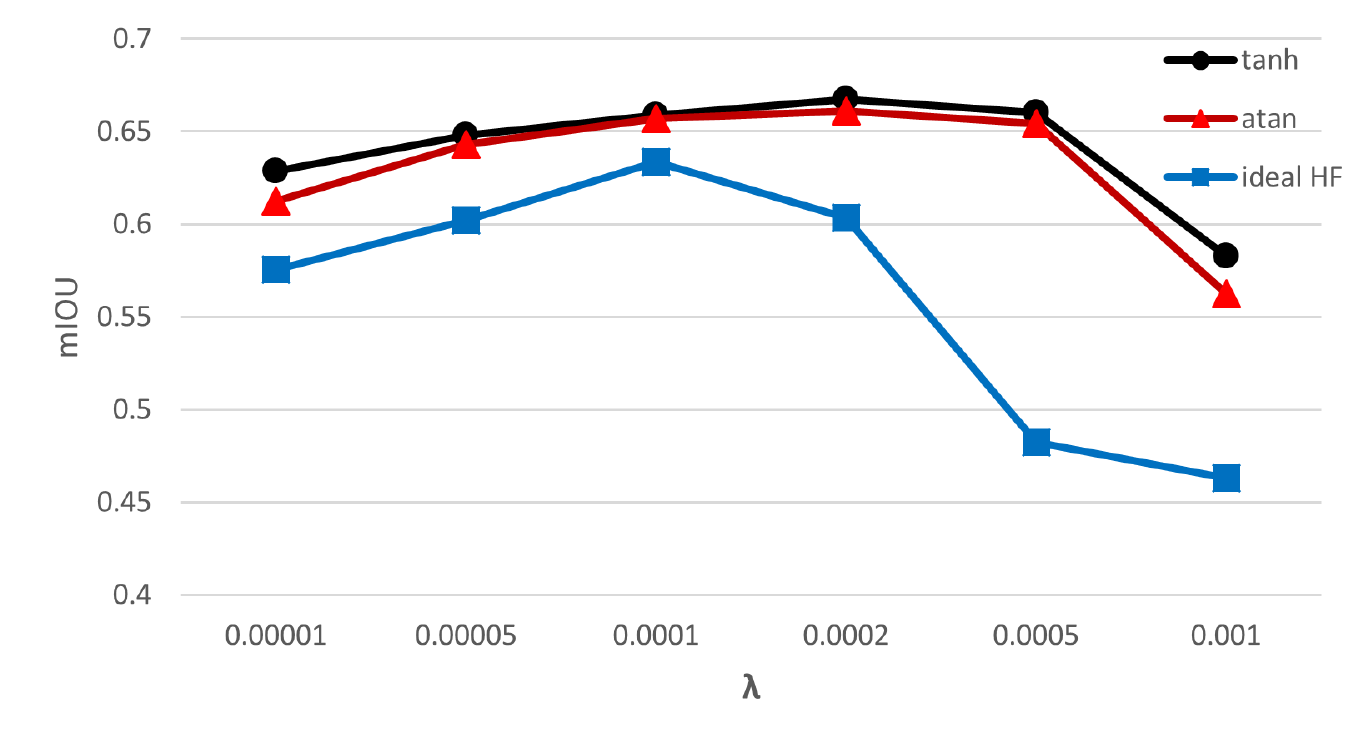}
	\caption{Comparison between  Modified Approximated Heaviside Function (MAHF : black),  Approximated Heaviside Function (AHF : red) and Heaviside Function (HF : blue).  DeepLab-LargeFOV and the PASCAL VOC validation set are used for the comparison. }
			\label{fig:atan}
\end{figure}

In our work, we propose the Modified Approximated Heaviside Function (MAHF) to apply the level set theory in deep learning.
We also compared the performance of MAHF with HF and AHF \cite{Levelset}.
For comparison, $\epsilon$ for AHF set to 20  so that the network shows the best performance.
Figure  \ref{fig:atan} shows the comparison between various Heaviside Function.
Note  that HF shows inferior performance since it is prone to be stuck in local minima.
MAHF and AHF achieve comparable performance, but MAHF gives slightly better results than AHF.

\section{Conclusion}
In this paper, we have proposed the level set loss for CNN-based semantic segmentation. 
Compared to the existing cross-entropy loss, our proposed loss considers spatial information of ground truth. 
The network trained with the level set loss represents the spatial information better and alleviates the typical problems of semantic segmentation.
Experimental results on representative networks, FCN and DeepLab, verify the generality of the loss.
Furthermore, unlike previous works, the level set loss does not require additional architectures, computational cost, and data.

For future work, we intend to develop loss functions that concern the spatial information of ground truth. 
The level set loss is just one of them. 
Our work notices the worth of designing appropriate losses to implement high performance segmentation networks. 

\vspace{2.8ex}
\hspace{-4ex}
\begin{large}
    \textbf{ACKNOWLEDGMENT}
\end{large}
\vspace{0.5ex}

This work was supported by the National Research Foundation of Korea (NRF) grant funded by the Korea government (MSIT) (No. 2018025409).

\begin{figure*}[p] \label{FCN}
	\centering
  	 	\begin{subfigure}[b]{0.18\textwidth}		
		\includegraphics[width=\textwidth]{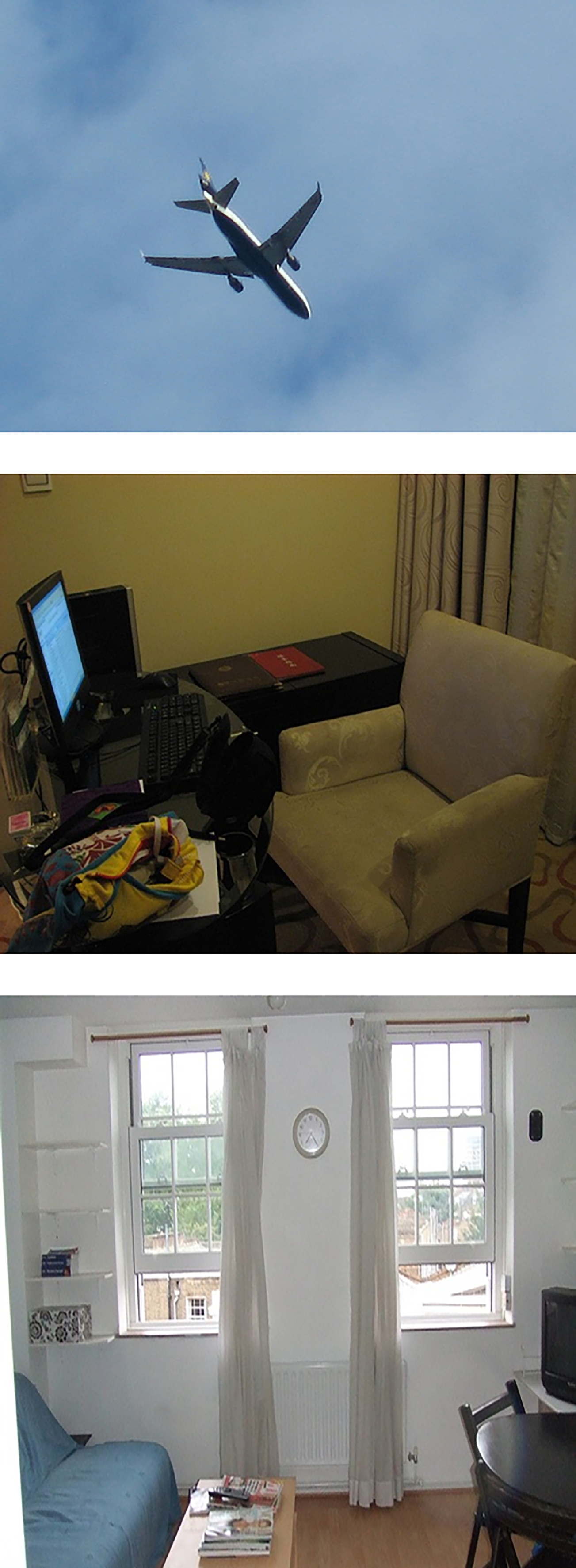}
		\caption{\scriptsize Image }
		\label{fig:quality_a}
	    \end{subfigure}
		\begin{subfigure}[b]{0.18\textwidth}		
		\includegraphics[width=\textwidth]{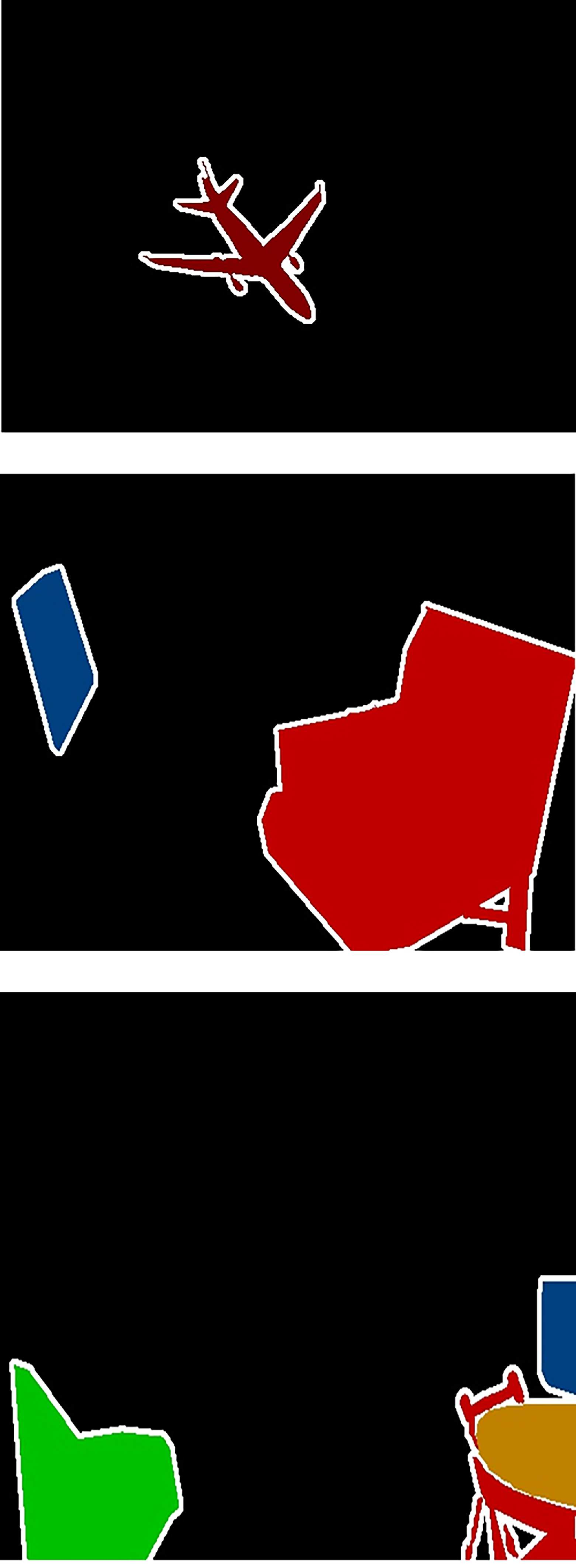}
		\caption{\scriptsize Ground \hspace{-1ex} Truth}
		\label{fig:quality_b}
	    \end{subfigure}
		\begin{subfigure}[b]{0.18\textwidth}		
		\includegraphics[width=\textwidth]{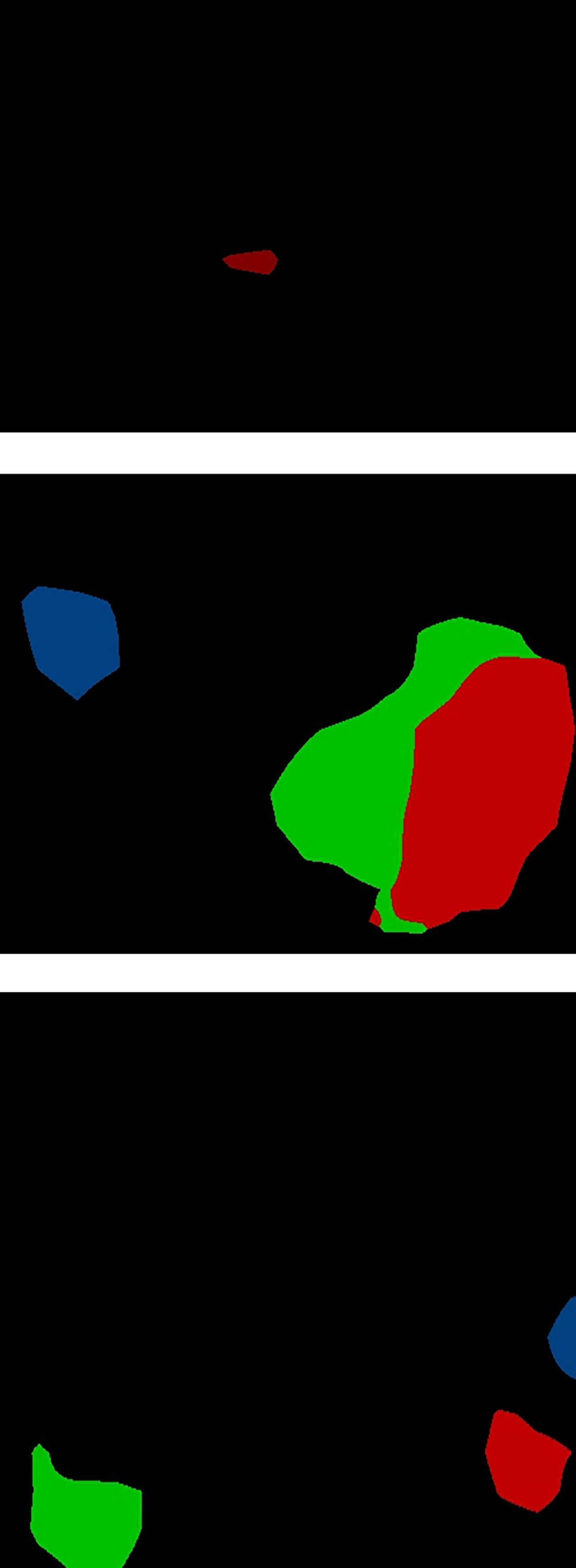}
		\caption{\scriptsize FCN 32s }
		\label{fig:quality_c}
	    \end{subfigure}
		\begin{subfigure}[b]{0.18\textwidth}		
		\includegraphics[width=\textwidth]{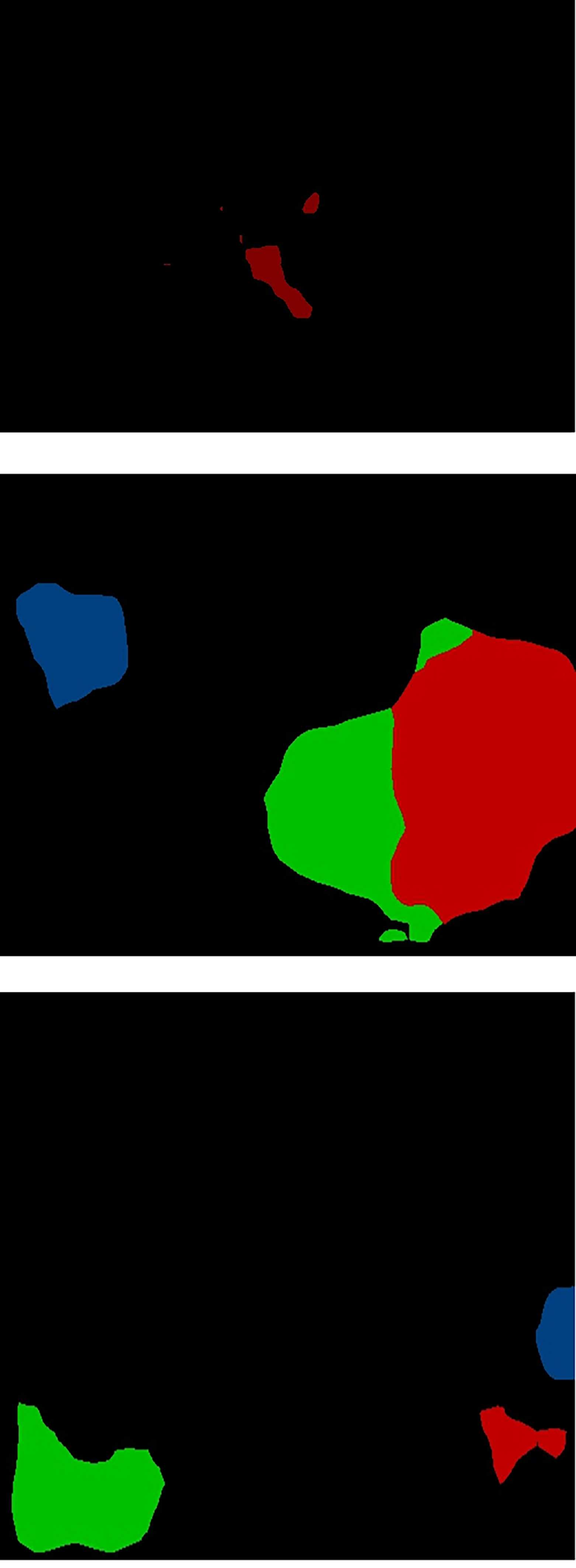}
		\caption{\scriptsize FCN 8s }
		\label{fig:quality_d}
	    \end{subfigure}
	    \begin{subfigure}[b]{0.18\textwidth}		
		\includegraphics[width=\textwidth]{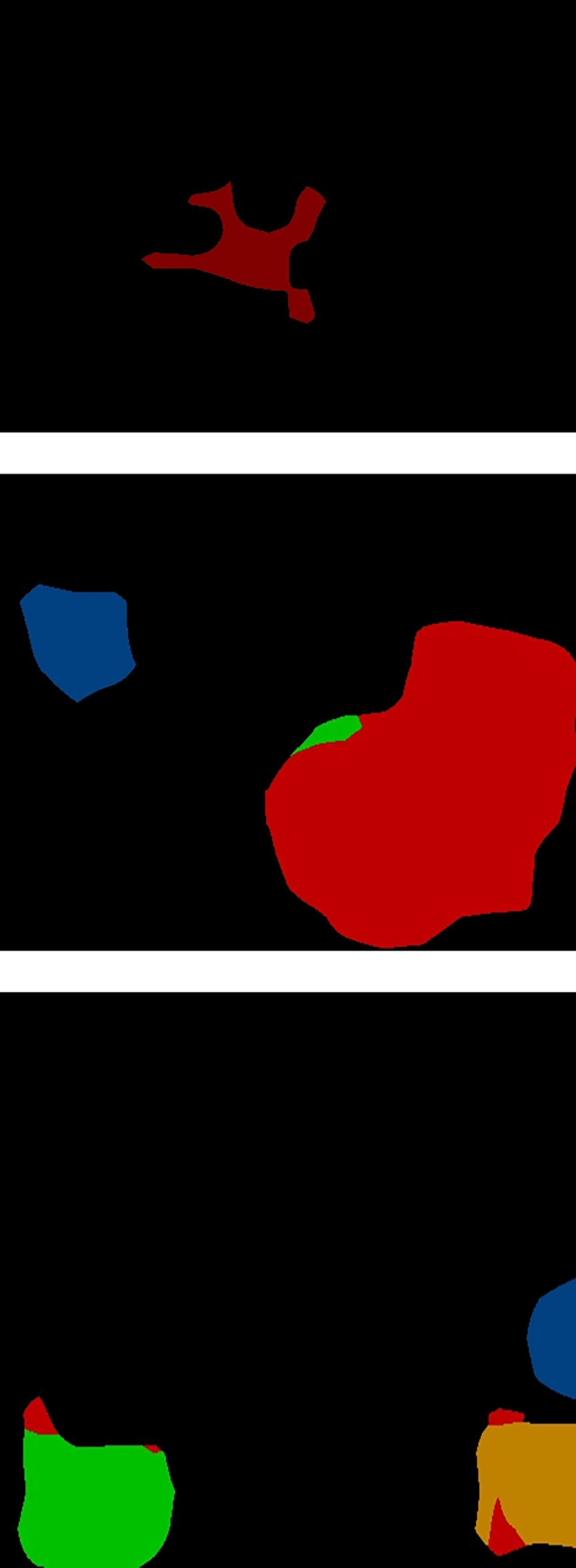}
		\caption{\scriptsize  32s + LS loss}
		\label{fig:quality_e}
	    \end{subfigure}
	\caption{Qualitative results of FCN-Resnet101 on the Pascal VOC 2012 $val$ set. Our proposed level set loss regularize the segmentation network  to represent global context information. The trained FCN 32s  with our level set loss shows better performance than FCN 8s.}
\end{figure*}

\begin{figure*}[p] \label{Deeplab}
	\centering
	\begin{subfigure}[b]{0.18\textwidth}		
		\includegraphics[width=\textwidth]{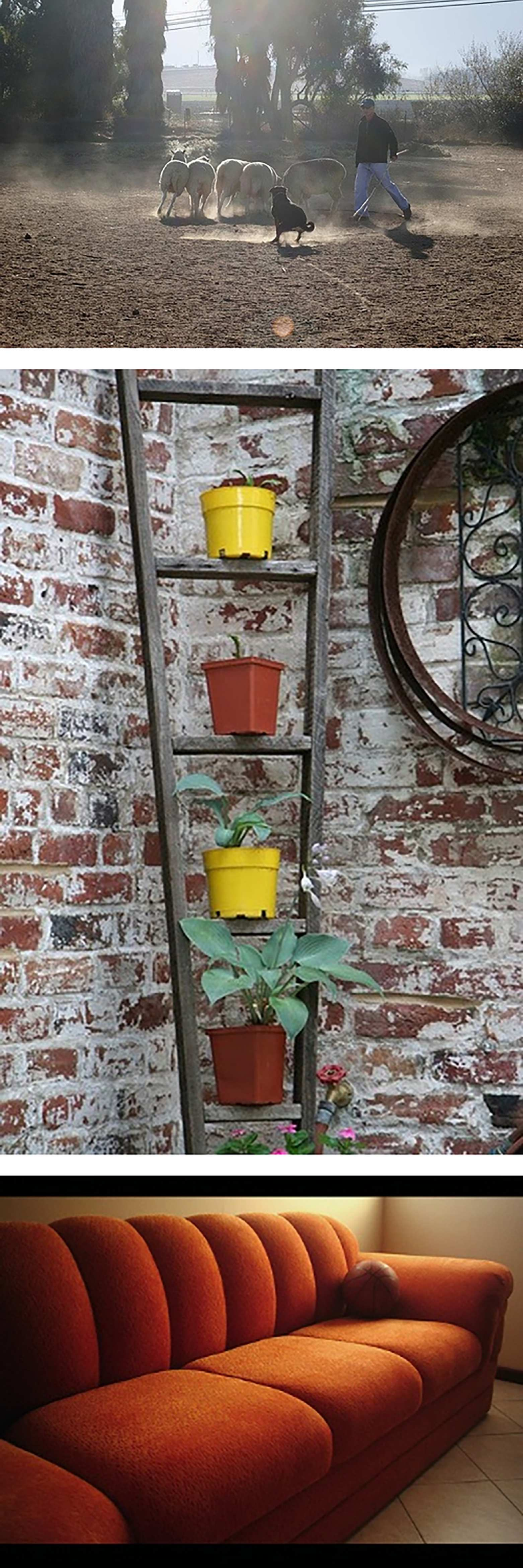}
		\caption{\scriptsize Image}
	\end{subfigure}
	\begin{subfigure}[b]{0.18\textwidth}	
		\includegraphics[width=\textwidth]{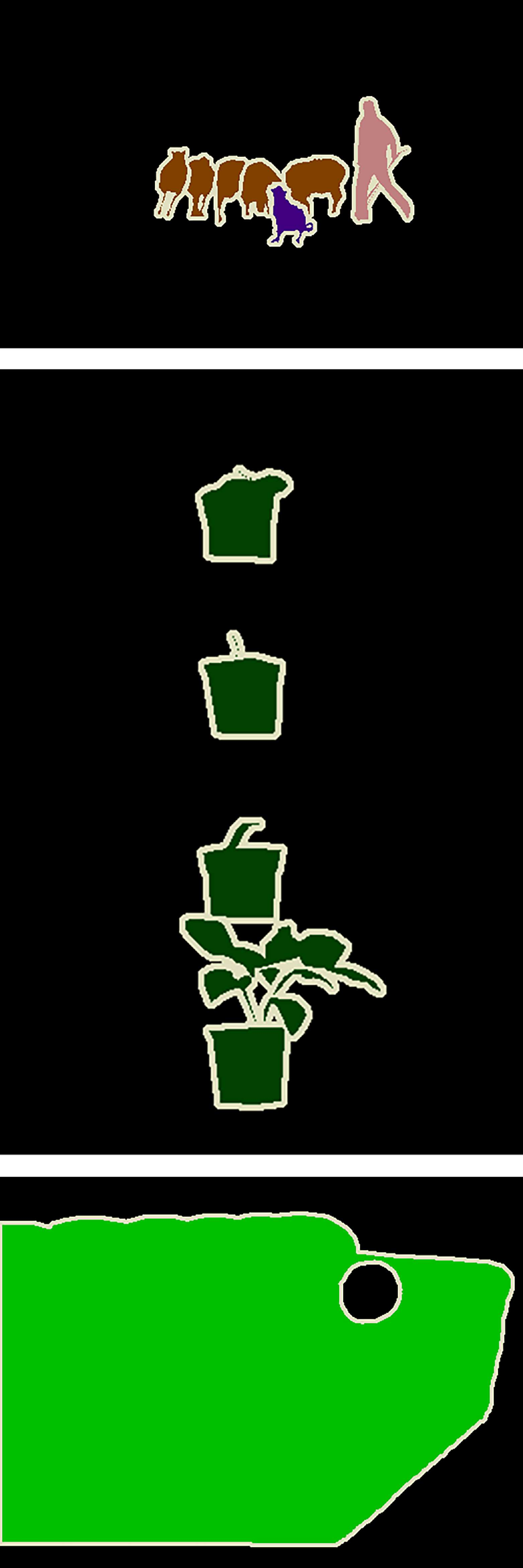}
		\caption{\scriptsize Ground \hspace{-1ex} Truth}
	\end{subfigure}
	\begin{subfigure}[b]{0.18\textwidth}
		\includegraphics[width=\textwidth]{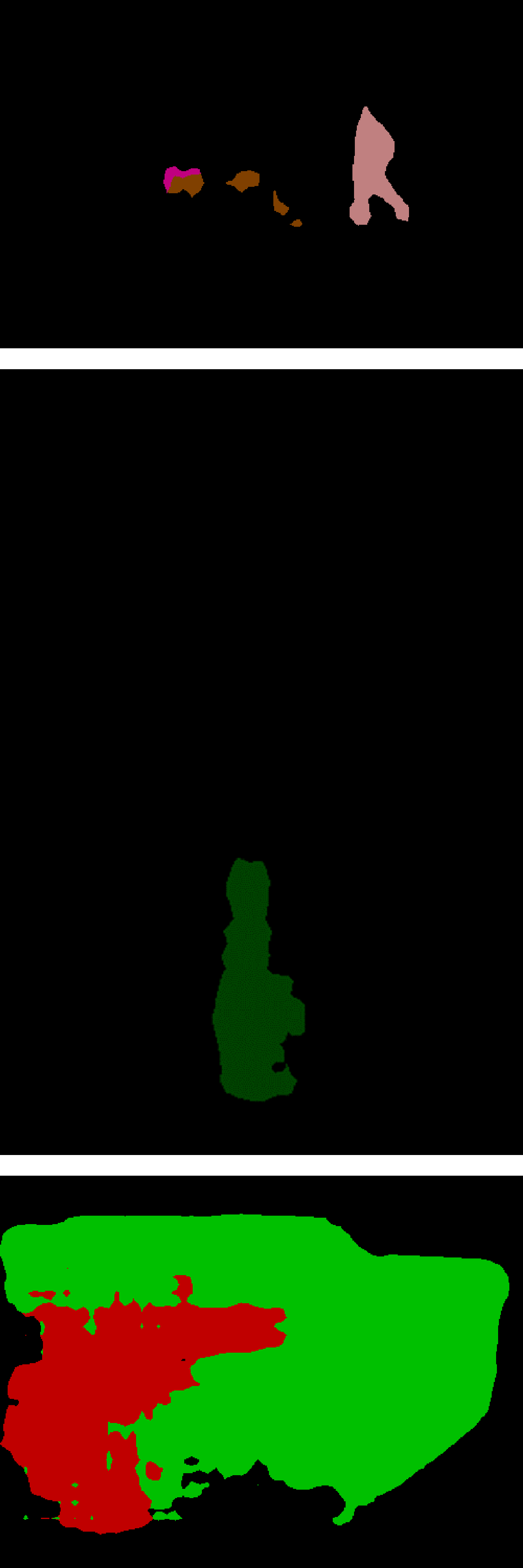}
		\caption{\scriptsize DeepLab}
	\end{subfigure}
	\begin{subfigure}[b]{0.18\textwidth}
		\includegraphics[width=\textwidth]{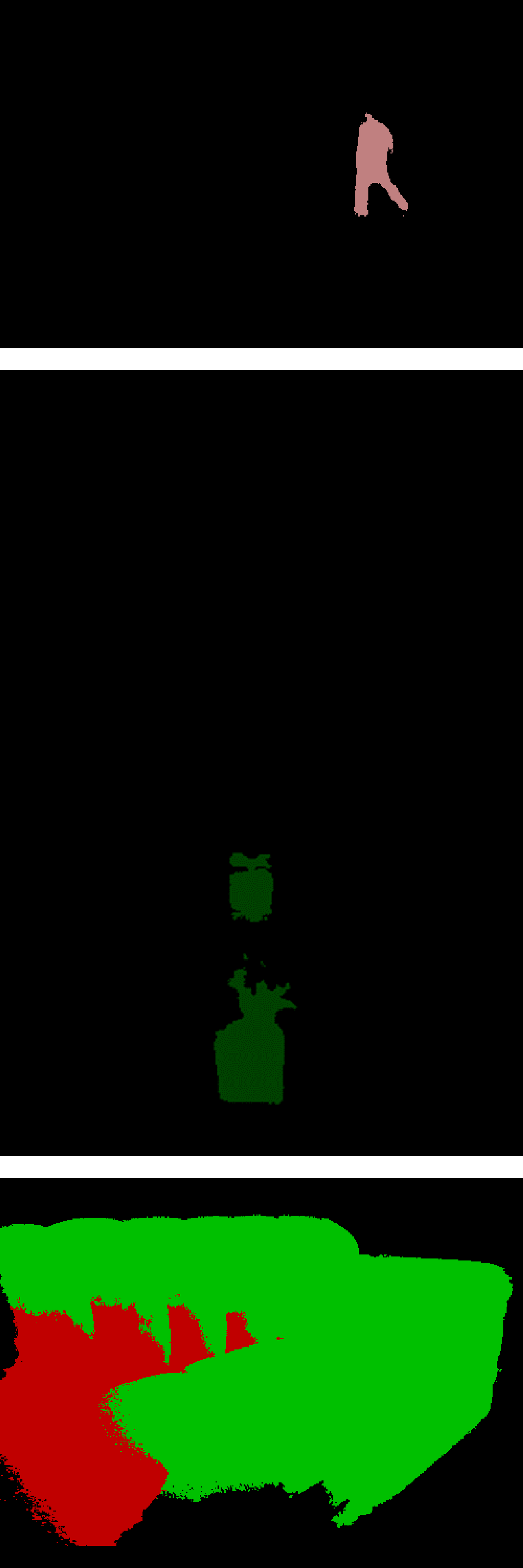}
		\caption{\scriptsize with CRFs}
	\end{subfigure}
	\begin{subfigure}[b]{0.18\textwidth}
		\includegraphics[width=\textwidth]{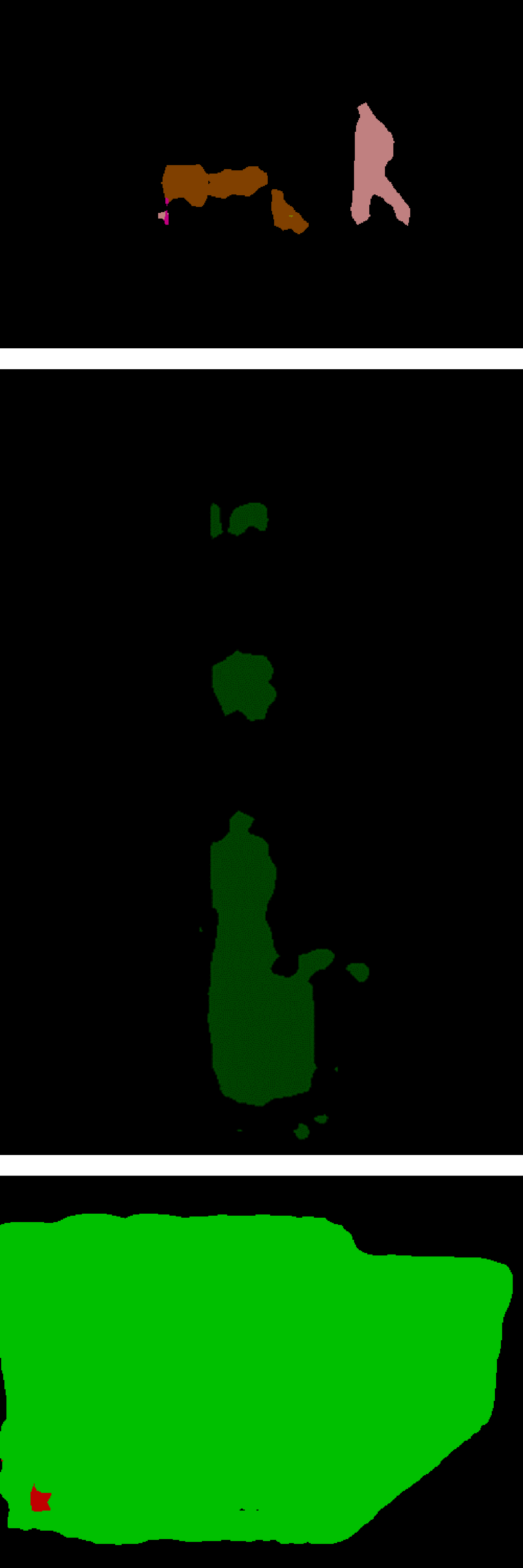}
		\caption{\scriptsize with LS loss}
	\end{subfigure}
	\caption{Qualitative results of DeepLab-ResNet101 on the Pascal VOC 2012 $val$ set. (d) is the result of using CRFs as post-processing. (e) is the results of the end-to-end training with our level set loss (No post-processing).}
		\label{qualitative_results}
\end{figure*}

\clearpage

{\small
\bibliographystyle{ieee}
\bibliography{egbib}
}

\end{document}